\documentclass[10pt,twocolumn,letterpaper]{article}

\usepackage{cvpr}
\usepackage{times}
\usepackage{epsfig}
\usepackage{graphicx}
\usepackage{amsmath}
\usepackage{amssymb}
\usepackage{subcaption}
\usepackage{makecell}
\usepackage{pbox}
\usepackage[titletoc]{appendix}
\usepackage{hyperref}

\newcommand{\cutsectionup}{\vspace*{-0.07in}}
\newcommand{\cutsectiondown}{\vspace*{-0.10in}}

\newcommand{\cutsubsectionup}{\vspace*{-0.07in}} 
\newcommand{\cutsubsectiondown}{\vspace*{-0.1in}} 

\newcommand{\cutcaptionup}{\vspace*{-0.1in}}
\newcommand{\cutcaptiondown}{\vspace*{-0.1in}}

\newcommand{\cuttableup}{\vspace*{-0.0in}}
\newcommand{\cuttabledown}{\vspace*{-0in}}

\newcommand{\hl}[1]{\textcolor{black}{#1}}

\newcommand{\pami}[1]{\textcolor{black}{#1}}

\cvprfinalcopy 

\def\cvprPaperID{641} 

\ifcvprfinal\fi
\begin{document}

\title{Recognition from Hand Cameras:\\ A Revisit with Deep Learning}

\author{Cheng-Sheng  Chan, Shou-Zhong Chen, Pei-Xuan Xie, Chiung-Chih Chang, Min Sun\\
National Tsing Hua University\\
{\tt\small \{s104061526@m104,s104061545@m104,s101061230@m101,s101060006@m101,sunmin@ee\}.nthu.edu.tw}}


\maketitle

\begin{abstract}
We revisit the study of a wrist-mounted camera system (referred to as HandCam) for recognizing activities of hands.
HandCam has two unique properties as compared to egocentric systems \cite{fathi2011learning,damen2014you} (referred to as HeadCam): (1) it avoids the need to detect hands; (2) it more consistently observes the activities of hands.
By taking advantage of these properties, 
we propose a deep-learning-based method to recognize hand states (free v.s. active hands, hand gestures, object categories), and discover object categories.
Moreover, we propose a novel two-streams deep network to further take advantage of both HandCam and HeadCam.
We have collected a new synchronized HandCam and HeadCam dataset with $20$ videos captured in three scenes for hand states recognition.
Experiments show that our HandCam system consistently outperforms a deep-learning-based HeadCam method (with estimated manipulation regions) and a dense-trajectory-based~\cite{wang:2011} HeadCam method in all tasks.
We also show that HandCam videos captured by different users can be easily aligned to improve free v.s. active recognition accuracy ($3.3\%$ improvement) in across-scenes use case.
Moreover, we observe that finetuning Convolutional Neural Network \cite{NIPS2012Alex} consistently improves accuracy.
Finally, our novel two-streams deep network combining HandCam and HeadCam features achieves the best performance in four out of five tasks.
With more data, we believe a joint HandCam and HeadCam system can robustly log hand states in daily life.
\end{abstract}


\cutsectionup
\section{Introduction}
\cutsectiondown

Recently, the technological advance of wearable devices has led to significant interests in recognizing human behaviors in daily life (i.e., uninstrumented environment). Among many devices, egocentric camera systems have drawn significant attention,
since the camera is aligned with the wearer's field-of-view, it naturally captures what a person sees.
These systems have shown great potential in recognizing daily activities (e.g., making meals, watching TV, etc.) \cite{pirsiavash2012detecting}, estimating hand poses \cite{RogezSKMR14,rogez2015first},  generating how-to videos \cite{damen2014you}, etc.


\begin{figure}[!t]
\includegraphics[width=0.5\textwidth]{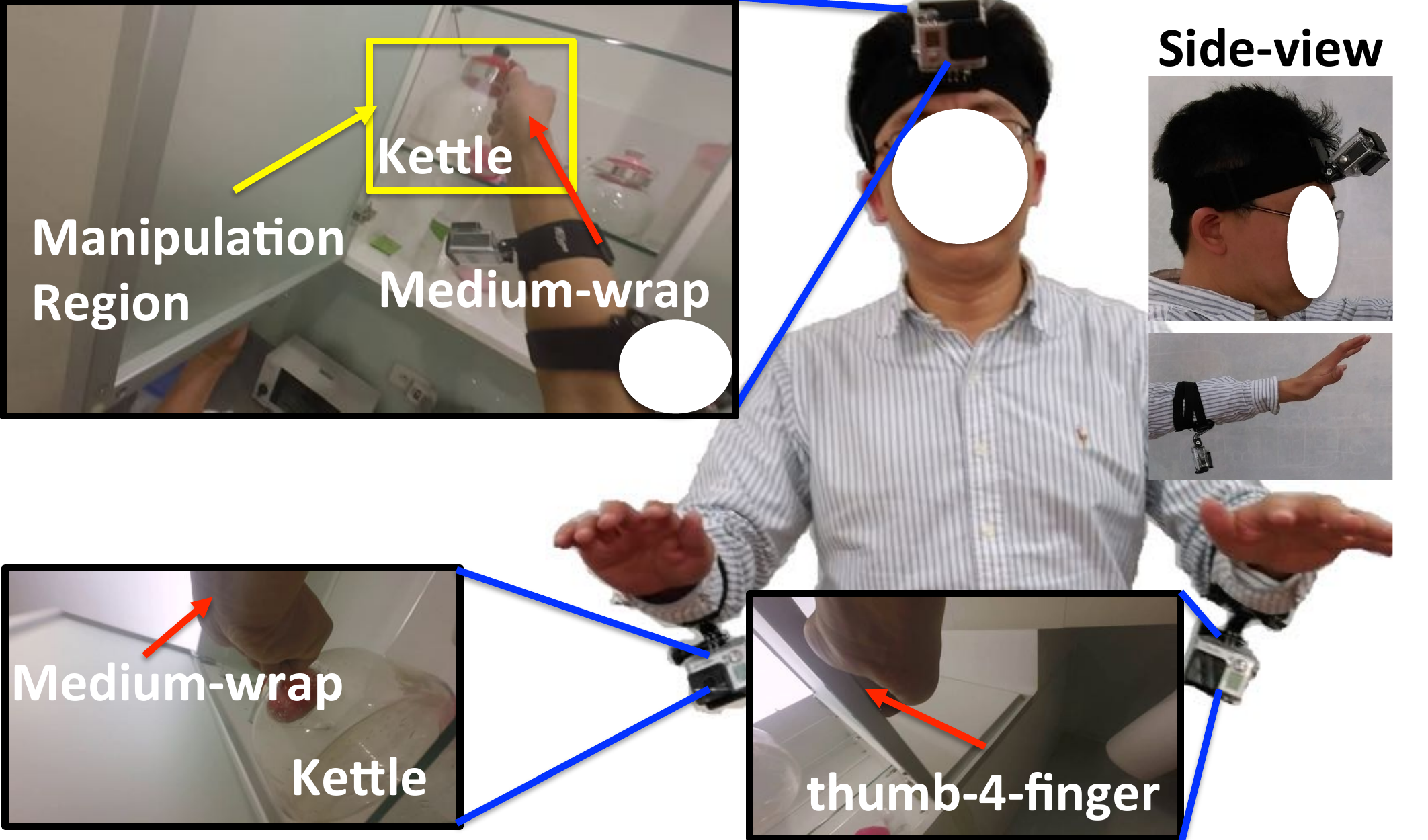}
\cutcaptionup
\caption{\small Illustration of our wearable camera system: consisting of three wide-angle cameras, two mounted on the left and right wrists to capture hands (referred to as HandCam) and one mounted on the head (referred to as HeadCam). We use our HandCam system to robustly recognize object categories (see yellow boxes for illustration) and hand gestures (see red arrows for illustration).}\label{fig.system}
\cuttabledown
\end{figure}

Despite many advantages of egocentric camera systems, there exists two main issues which are much less discussed \cite{RogezSKMR14}.
Firstly, hand localization is not solved especially for passive camera systems.
Even for active camera systems like Kinect, hand localization is challenging when two hands are interacting or a hand is interacting with an object. Secondly, the limited field-of-view of an egocentric camera implies that hands will inevitably move outside the images sometimes.
On the other hand, cameras have been mounted on other locations to avoid similar issues.
In project Digit \cite{KimDigit}, a camera is mounted on a user's wrist to always observe the user's hand pose. This allows a user to issue gesture commands at any time.
Similarly, a camera has been mounted on a robot arm for accurately picking up an object~\cite{Saxena:2008grasp}.
In fact, the seminal work \cite{Placement} has conducted simulation on 3D model of the human body to analyze the effects of field of view and body motion, when cameras are mounted at different locations.
Hence, we argue that egocentric camera system might not be the best wearable system for recognizing human behaviors.

We revisit the wrist-mounted camera system (similar to \cite{Maekawa10}) to capture activities of both hands (Fig.~\ref{fig.system}). We name our system ``HandCam" which is very different from egocentric systems with cameras on head or chest (e.g., \cite{fathi2011learning,damen2014you,RogezSKMR14}). By wearing cameras on wrists, HandCam directly recognizes the states of hands (e.g.,  object: kettle; gesture: medium-wrap in Fig.~\ref{fig.system}). It avoids the needs to detect hands and infer manipulation regions as required in classical egocentric systems \cite{fathi2011learning}.
\pami{A few methods have been proposed to recognize activities using wrist-mounted camera~\cite{Maekawa10,maekawa2012wristsense}.
They show that wrist-mounted sensor system can be small and user-friendly. They also primarily focus on fusing different sensing modalities.
However, we focus on designing a deep-learning-based vision algorithm to
improve recognition accuracy (see Sec.~\ref{sec.rwh} for more comparison).}
Most importantly, we are one of the first to propose a novel two-streams deep network taking advantages of both HandCam and HeadCam.
All our methods are design to classify hand states including free v.s. active (i.e., hands holding objects or not), object categories, and hand gestures (Sec.~\ref{sec.ustate}).
A similar method is also proposed to discover object categories in an unseen scene (Sec.~\ref{sec.DC}).


To evaluate our system, we collected a new synchronized HandCam and HeadCam dataset for hand state recognition.
The dataset consists of 20 sets of video sequences (i.e., each set includes two HandCams and one HeadCam synchronized videos) captured in three scenes: a small office, a mid-size lab, and a large home.
In order to thoroughly analyze recognition tasks, we ask users to interact with multiple object categories and multiple object instances. We also ask multiple users to wear HandCam in a casual way to consider the variation introduced by multiple users. To overcome this variation, a fully automatic hand alignment method is proposed (Sec.~\ref{sec.A}).

Experiments show that our HandCam system consistently outperforms
a deep-learning-based HeadCam method (with estimated manipulation regions~\cite{li2013learning}) and a dense-trajectory-based~\cite{wang:2011} HeadCam method in all tasks.
Moreover, we show that HandCam videos captured by different users can be easily aligned to improve free v.s. active recognition accuracy ($3.3\%$ acc. improvement) in across-scenes use case.
In all experiments, we use state-of-the-art Convolutional Neural Network (CNN)\cite{NIPS2012Alex} features.
We observe that finetuning CNN consistently improves accuracy (at most $4.9\%$ improvement). 
Finally, our method combining HandCam and HeadCam features achieves the best performance.

\pami{A preliminary version of the paper is under submission. In this paper, we include an additional HandCam baseline using a strong but time-consuming to compute dense trajectory feature. We also include detail experiments on exploring different CNN architectures, more visualization of the estimated states in a videos, many typical examples of hand states recognition, and an error analysis to identify potential direction for future improvement.}
\cutsectionup
\section{Related Work}
\cutsectiondown

A few non-vision-based methods have been proposed to recognize human daily activities based on recognizing hand states~\cite{Wrist05,Stikic08}.
Wu et al~\cite{WuICCV07} combine sparse RFID data with a third-person video to recognize human daily activities based on objects used.
In the following, we focus on the vision-based methods and review related work in egocentric recognition, hand detection and pose estimation, and a few works inspired our HandCam system.

\cutsubsectionup
\subsection{Egocentric Recognition}
\cutsubsectiondown
\vspace{1mm}

\begin{figure*}[!t]\cuttableup
\centering
\includegraphics[width=1\textwidth]{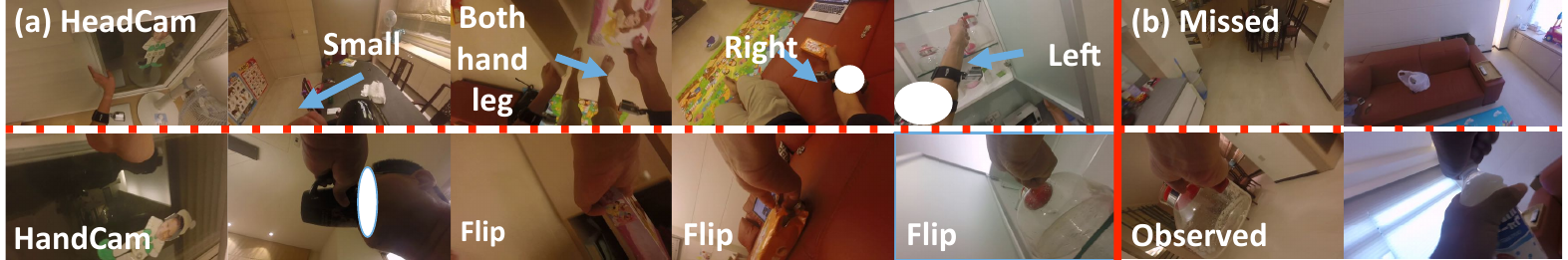}
\cutcaptionup
\caption{\small HandCam (bottom-row) v.s. HeadCam (top-row). Panel (a) compares the hand location variation. The variation in HandCam (bottom) is significantly less than variation in HeadCam (top).  We also know exactly which video captures left or right hand. We flip left hand images to mimic right hand images and train a single deep network.
Panel (b) shows typical examples of missed hands in HeadCam but observed hands in HandCam. For example, we do not require consistent hand-eye coordination for opening a water bottle while walking.
}\label{fig.HandvsHead}
\cuttabledown
\end{figure*}

\cite{fathi2011learning,fathi2011understanding,fathi2013State} are the early egocentric works learning to recognize objects, actions, and activities.
These methods assume foreground objects and hands can be easily separated from background using appearance, geometric, and motion cues.
Their methods are evaluated on an egocentric activity dataset where the users move mainly their hands in front of a static table.
In contrast, we allow users to naturally move to different places in a scene, which creates addition challenge for egocentric system to localize hands.
\hl{For instance, Pirsiavash and Ramanan~\cite{pirsiavash2012detecting} also propose to recognize activities through recognizing objects while users is moving in their homes.
Since their approach is based on detecting hand-object manipulation in the egocentric field-of-view,
it is confused mainly with activities observing similar objects in the field-of-view without manipulation.
However, since our HandCam significantly reduces hand location variation (Fig.~\ref{fig.HandvsHead}(a)), this scenario won't be a big issue for our HandCam system.}


\cite{fathi2012learning,li2013learning,li2015delving} further show the importance of gaze to help recognizing actions requiring ``hand-eye coordination". We argue that not all daily activities consistently requires hand-eye coordinate. For instance, we do not require consistent hand-eye coordination for opening a water bottle while walking (Fig.~\ref{fig.HandvsHead}-b-Right). In such case, head movement and gaze information might be misleading, and the user's hand and object of interest might move outside the field-of-view of the egocentric camera (Fig.~\ref{fig.HandvsHead}(b)). On the other hand, our HandCam system more consistently captures hand-object interaction for a variety of daily activities.
Finally, a few works~\cite{ghosh2012discovering,lu2013story,SMECCV14} focus on summarizing egocentric videos by recognizing objects, people, and scenes.



\noindent\textbf{Extra sensors.}
Fernando et al.~\cite{DeLaTorreHMV09} utilize motion capture techniques, static cameras, wearable IMUs, and a head-mounted camera to study food preparation process in an instrumented kitchen.
\cite{damen2014you} proposes to combine egocentric camera with gaze tracker to robustly ``discover" objects of interest given multiple sequences recorded in the same scene conducting the same tasks by different subjects. 
Moghimi et al.~\cite{501} propose to use a head-mounted RGBD camera to recognize both manipulation and non-manipulation activities.
Damen et al.~\cite{damen2012egocentric} also propose to use RGBD camera to model background and ``discover" foreground objects. 
In this work, we show that with our HandCam system, objects can also be discovered without the need of additional gaze tracker or RGBD sensor (Sec.~\ref{sec.DC}).

\cutsubsectionup
\subsection{Hand Detection and Pose Estimation}\cutsubsectiondown
\cite{RogezSKMR14,rogez2015first} focus on estimating 3D hand poses using wearable RGBD camera. Despite many success in 3D hand pose recognition, Rogez et al.~\cite{RogezSKMR14} show that egocentric 3D hand poses estimation is very challenging due to common interaction between hands and other objects or scene elements.
\cite{li2013pixel,li2013model} focus on detecting hand pixels in RGB images while users are moving around various environments.
Betancourt et al.~\cite{HandCVPRW14} study the weakness of \cite{li2013pixel} and proposes method for reducing false positive detection of hands. Although these RGB methods are not as robust as RGBD methods, these methods have been applied to discover hand gestures \cite{De2015}.


\cutsubsectionup
\subsection{Camera for Hands}\label{sec.rwh}
\cutsubsectiondown
A few work have proposed to wear cameras on wrists or other body parts to recognize gestures, poses, and activities of hands.
In \cite{KimDigit,Wrist99}, cameras are mounted on a user's wrists to always observe user's hand pose. This allows a user to issue gesture commands at any time. However, the project assumes that a user is hand free of objects. In contrast, our HandCam system focuses on recognizing hand-object interactions.
Similarly, a camera has been mounted on a robot arm for it to accurately pick up an object~\cite{Saxena:2008grasp}.
Although the robot has other sensors like a stereo camera and a lazer range finder which are not mounted on the robot arm, it has been shown that the HandCam is essential for picking up an object.
Chan et al.~\cite{CyclopsRing15} recently propose to wear a camera on hand webbings to recognize hand gestures and context-aware interactions such as events triggered by object recognition.
However, they assume that objects are instrumented with QR codes.
These works suggest that egocentric camera system might not be the best wearable system for understanding human behaviors.

\pami{Maekawa et al.~\cite{Maekawa10} is the most relevant prior work aiming for object-based activity recognition using sensors on wrist including a camera.
We share the same idea to take advantage of wrist-mounted camera to recognize human-object interaction.
However, it focuses on fusing the observation of heterogeneous sensors including a camera, a microphone, and an accelerometer.
Compared to our deep learning approach, they utilize simple and efficient color histogram as the feature.
Moreover, they train/test in the same environment and use same object instances, whereas we train/test different object instances across different environments (e.g., train: lab+office; test: home).
Ohnishi et al.~\cite{OhnishiKKH15Wrist} present a recent paper that achieves an outstanding recognition accuracy using a wrist-mounted camera (only on right hand).
They also focus on vision approach using deep features and dense-trajectory-based features~\cite{wang:2011}.
Our HandCam method does not use motion feature~\cite{wang:2011}, since it is time consuming to compute (on average a few seconds for each prediction).
On the other hand, our deep feature can be computed in real-time on a GPU.
Moreover, they use pre-trained CNN feature only, whereas we train a novel two-streams CNN to explore the advantage on learning representation for wearable cameras.
Finally, in their experiments, they assume that the temporal segment of each action is given. Hence, they evaluate per-segment classification accuracy,
whereas we evaluate per-frame classification accuracy.
}

\cutsectionup
\section{Our System}
\cutsectiondown

Our wearable camera system (Fig.~\ref{fig.system}) consists of three wide-angle cameras: two HandCams and one HeadCam.
We first propose a method utilizing deep-learning techniques to classify hand states (free v.s. active, object categories, and hand gestures) observed by either HandCam or HeadCam separately.
Finally, we propose a novel two-streams CNN model to take advantage of both HandCam and HeadCam in Sec.~\ref{sec.comb}.

\cutsubsectionup
\subsection{Wearable Cues}
\cutsubsectiondown

The strength of a wearable system essentially lies in the unique cues it can extract.
For an egocentric system, these cues include gaze, hand, and foreground information.
Some systems utilize active sensors to reliably localize hands (e.g., RGBD sensor in \cite{rogez2015first}) or predict user's attention (e.g., gaze tracker in \cite{damen2014you}). However, they require extra hardware expenses and more power consumption.
Many other egocentric systems require only a camera. However, sophisticated pre-processing steps \cite{fathi2011learning,fathi2011understanding,li2015delving,li2013learning} are required for removing irrelevant information in background.


Our HandCam system is designed with two focuses:
\vspace{-2mm}
\begin{itemize}
\item{Stable Hand Cue.} Significantly reduced hand location variation (Fig.~\ref{fig.HandvsHead}(a)-bottom) as compared to egocentric systems which have larger hand location variation (Fig.~\ref{fig.HandvsHead}(a)-top). Our system also won't be confused between left and right hands, since they are recorded by different cameras. Instead, we augment our dataset by flipping left hand images to mimic right hand images.
\pami{This can be considered as a data augmentation procedure~\cite{WuYSDS15} commonly used for training deep networks.}
\item{Consistent Observation.} Almost all hand related activities are observed as compared to egocentric systems which have limited field-of-view that missed some hand related activities (Fig.~\ref{fig.HandvsHead}(b)). 
\end{itemize}

\begin{figure}[!b]
\centering
\includegraphics[width=0.48\textwidth]{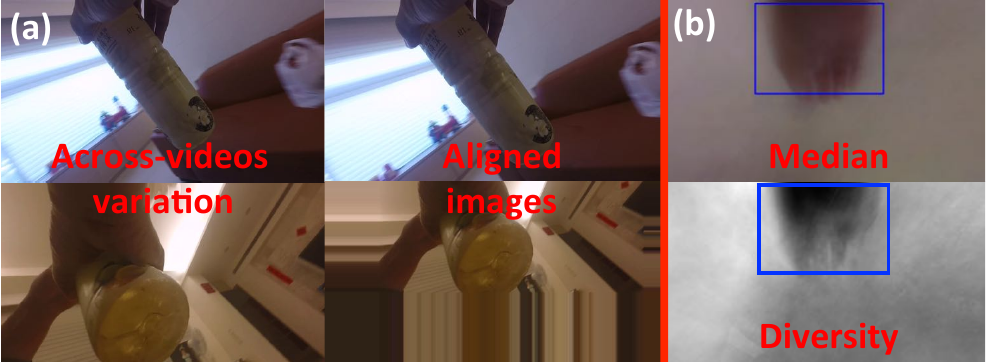}
\cutcaptionup
\caption{\small Across-videos Hand Alignment. Panel (a)-left shows the across-videos hand variation. 
Panel (a)-right shows aligned images.
Panel (b) shows example of median and diversity images on the top and bottom, respectively.}\label{fig.align}
\end{figure}

\noindent\textbf{Human factors.}
\pami{As we design our system to be used by general users, we let each user to wear the HandCam under a general guideline.
As a consequence, different users mount the HandCam with slightly different distances and orientation.}
Therefore, the hand location variation ``across" video sequences (Fig.~\ref{fig.align}(a)-left) is noticeable.
However, once the camera is mounted on a user's wrists, it is true that the spatial variation of hand regions are small within the video. 
By utilizing this fact, we propose a fully automatic across-videos hand alignment method.

\cutsubsectionup
\subsection{Hand Alignment}\label{sec.A}
\cutsubsectiondown


In each video sequence, we model the pixel value (i.e., a value between $0\sim 255$) distribution across time for each pixel and each color channel as a Laplace distribution parameterized by its center ($\mu$) and its diversity ($\beta$).  We estimate the parameters of the distribution using maximum likelihood estimators,
where the median image represents the common pattern (Fig.~\ref{fig.align}(b)-top) and the diversity image represents the variation of pixel values (black indicates small variation in Fig.~\ref{fig.align}(b)-bottom). We simply treat the regions with diversity $\beta$ smaller than $\beta_{th}$ for all color channel as ``stable" hand mask (within blue box in Fig.~\ref{fig.align}(b)-bottom). We find the video with the smallest ``stable" hand mask region as the reference video, and use the median of the region as alignment template  (blue box in Fig.~\ref{fig.align}(b)-top) . 
\hl{We apply multiscale normalized cross-correlation to align the template to the median images of other videos.
 Then, we apply cropping and replicate padding to generate hand aligned images (Fig.~\ref{fig.align}(a)-right).
}

%

\cutsubsectionup
\subsection{Hand States Recognition}\label{sec.ustate}
\cutsubsectiondown

\begin{figure}[!b]\cuttableup
\centering
\includegraphics[width=0.48\textwidth]{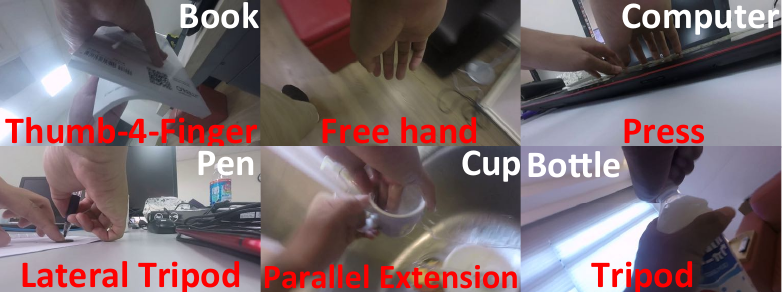}
\cutcaptionup
\caption{\small Typical hand states in HandCam view: object category (top-white-font) and hand gesture (bottom-red-font). The statisitcs of states in our dataset in shown in Table.~\ref{fig:stat}}\label{fig.states}
\cuttabledown
\end{figure}

Given the aligned (Fig.~\ref{fig.align}(a)-right) and stable  (Fig.~\ref{fig.HandvsHead}(b)-bottom) observation of hand, we propose to recognize the hand states for multiple tasks (Fig.~\ref{fig.states}).

\noindent\textbf{Free v.s. active.}
The most fundamental states of interests is to recognize whether the hand is manipulating an object (referred to as ``active state"), or not  (referred to as ``free state").
In this work, we explicitly evaluate the performance of active hands recognition undergoing some unseen activities.

\begin{figure}[!t]\cuttableup
\centering
\includegraphics[width=0.48\textwidth]{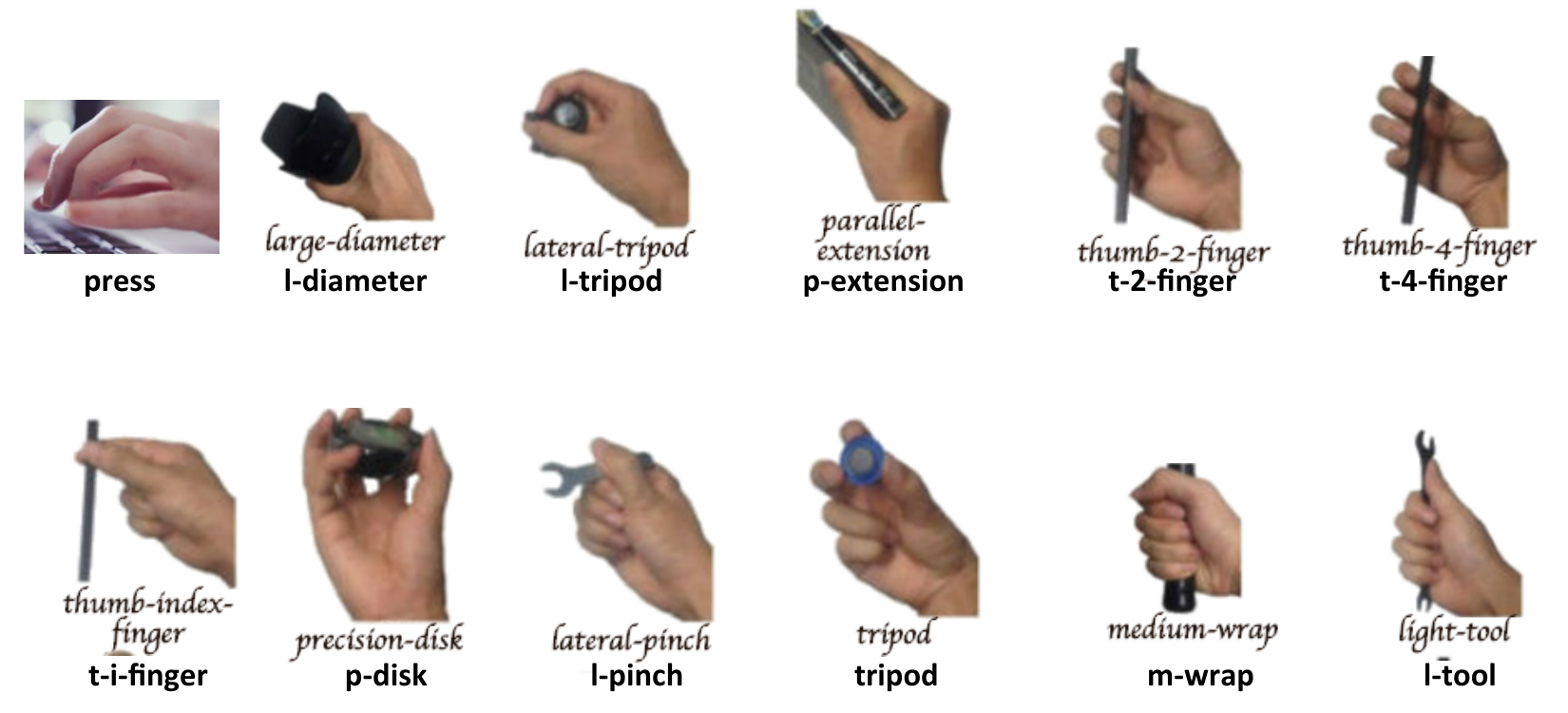}
 \cutcaptionup
  \caption{\small Visualization of 12 gesture classes. Except ``press", visualization of all other gestures are reused from \cite{De2015}.}
  \label{fig:ges}
  \cutcaptiondown
\end{figure}

\noindent\textbf{Hand gesture recognition.}
At a slightly finer level of granularity (12 gesture classes shown in Fig.~\ref{fig:ges}), we propose to recognize hand gestures of active hands. Note that gesture is an important affordance cue for recognizing activities without the need to recognize a large number of object categories involving in the same activity. 

\noindent\textbf{Object category recognition.}
At the finest level of granularity (23 object categories), we propose to recognize object categories which have been manipulated by hands. Categorical recognition allows our system to recognize an unseen object instance within a know category. 



We take a fully supervised approach and train a frame-based multiclass state classifier.
The classifier generates a confidence $u(s_i)$ for state $s_i$ in the $i^{th}$ frame.
\pami{For example, $u(s_i=\textit{Active})$ specifies the confidence that the $i^{th}$ frame contains an active hand.
$u(s_i=\textit{Notebook})$ specifies the confidence that the $i^{th}$ frame contains a hand manipulating a notebook.
We take advantage of the recent breakthrough in deep learning and extract frame-based feature $f_i$ from Convolutional Neural Network (CNN) \cite{NIPS2012Alex}, where $i$ denotes the frame index.
The deep feature $f_i$ is a high-level representation used as the input of the state classifier and the state change detector described next.}
We describe the setting of the CNN model that we use in our application in Sec.~\ref{sec.Dmodel} and Sec.~\ref{sec.comb}.



\cutsubsectionup
\subsection{State Change Detection}\label{sec.bstate}
\cutsubsectiondown

Since frame-based state recognition inevitably will contain spurious error predictions,
we propose to restrict the locations of possible state changes by learning to detect them.

\begin{figure}[!t]\cuttableup
\centering
\includegraphics[width=0.48\textwidth]{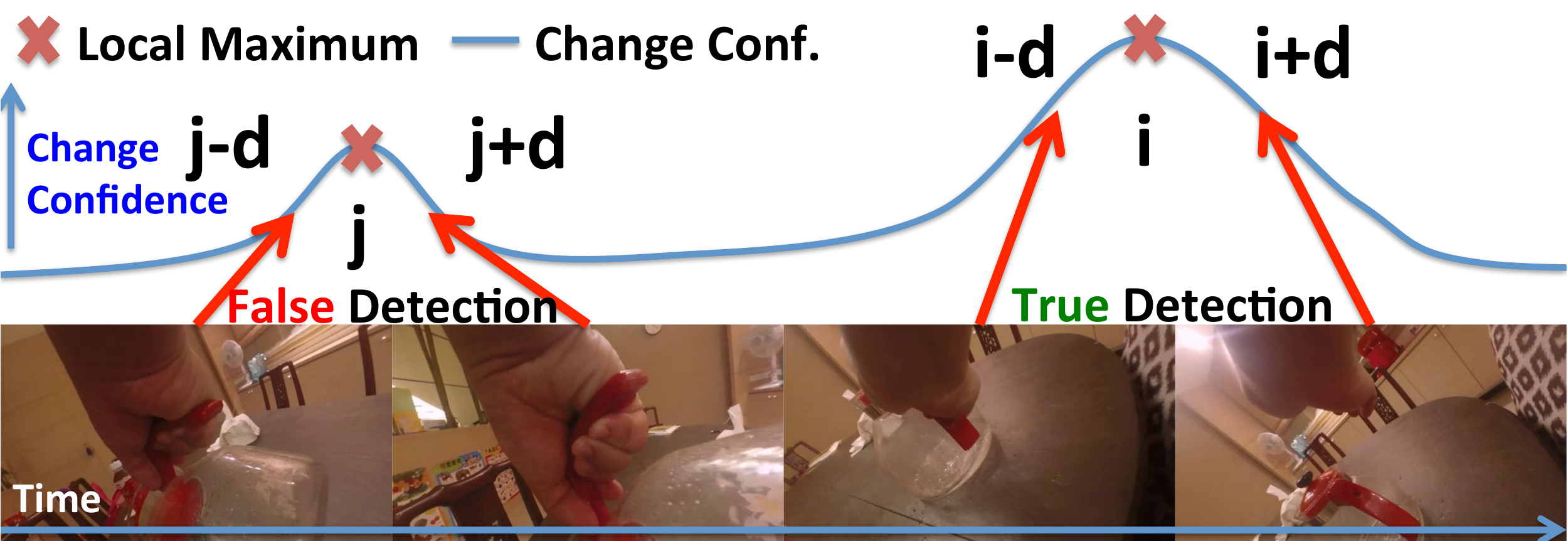}
\cutcaptionup
\caption{\small Illustration of state change detection. Two change candidates are shown, where the first one corresponds to a false detection and the second one corresponds to a true detection. Our system ensures high recall of state changes at this step.}\label{fig.Ch}
\cuttabledown
\end{figure}

\noindent\textbf{Frame-based change.}
We train a frame-base binary state change classifier (i.e., change or no change) by treating frames within $d$ frames distance away from a ground truth change frame as positive examples and remaining frames as negative examples. The large value of $d$ will increase the number of positive examples, but decreasing the localization ability of the true change locations.
In order to reduce the visual variation of changes, we propose the following feature,
\begin{eqnarray}
cf_i = |f_{i-d}-f_{i+d}|= |f_{i+d}-f_{i-d}|~,
\end{eqnarray}
\pami{where $f$ is the same deep feature used for state classifier.
Note that $f$ is a high-level semantic feature (not a low-level pixel-wise color or intensity feature).
Hence, $cf$ measures semantic changes, but not low-level motion or lighting condition changes.
Moreover, $cf$ implies that transition from active to free should have a similar feature representation as transition from free to active.
Given $cf$, we apply a change classifier to obtain frame-based state change confidences for all frames (Fig.~\ref{fig.Ch}).}

\noindent\textbf{Change candidates.}
Similar to classical edge detection \cite{Canny}, we need to remove redundant change candidates with high confidences.
Hence, we apply non-maximum suppression to find local maximum with respect to state change confidences (Fig.~\ref{fig.Ch}).
We define the local maximum locations as change candidates $\{i\}_{i\in C}$, where $C$ contains a set of local maximum change locations.
Note that we prefer high recall (i.e., allowing some false candidates in Fig.~\ref{fig.Ch}) at this step.


\begin{figure*}[!t]\cuttableup
\centering
 \includegraphics[width=0.21\textwidth]{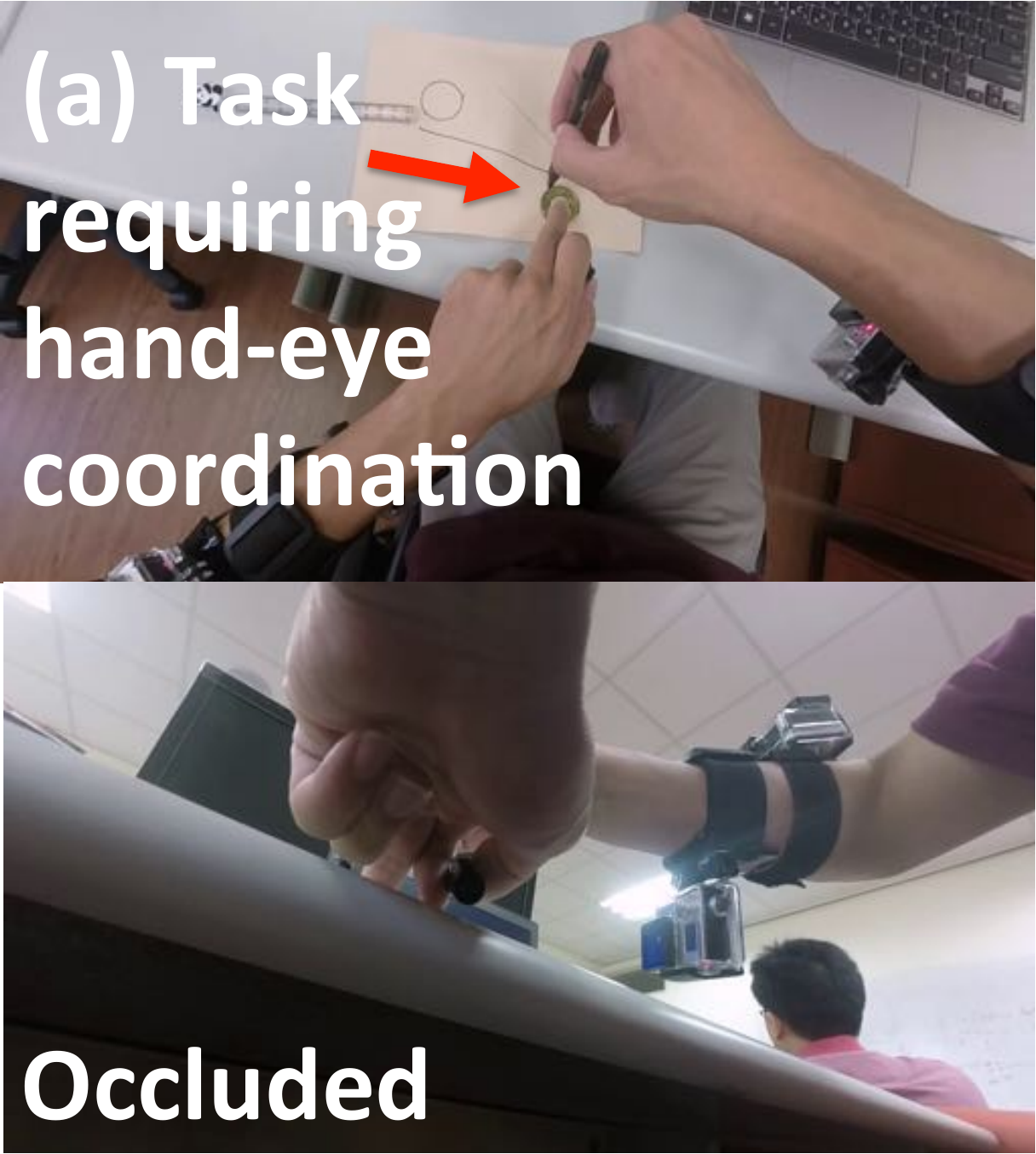}
 \includegraphics[width=0.75\textwidth]{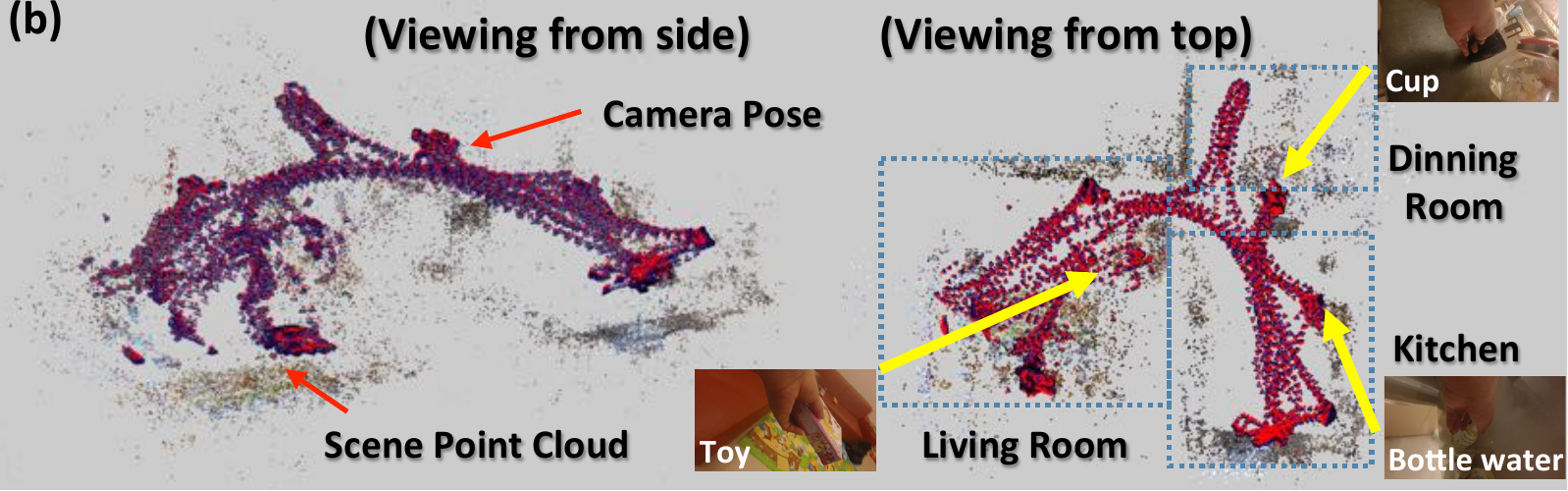}\\
\vspace{4mm}
\caption{\small Panel (a) shows an example where HandCam (bottom) is occluded but HeadCam (top) observed the activity requiring hand-eye coordination. Panel (b) shows 3D Structure~\cite{Changchang13DV} of the Scene reconstructed from HeadCam images. A pair of blue and red dots indicates the recovered camera 3D pose and other color-coded dots show the color of the scene. This shows that HeadCam contains place information which potentially is useful for hand states recognition.
}\label{fig.3D}
\cuttabledown
\end{figure*}

\cutsubsectionup
\subsection{Full Model}\label{sec.F}
\cutsubsectiondown
We now combine frame-based state classification with detected change candidates to improve the classification of states.
Both information are captured into a pairwise scoring function below,
\begin{eqnarray}
R(S) = \sum_{i=1}^{N}u(s_i)+\lambda \sum_{i=1}^{N-1}b(s_i,s_{i+1})~,
\end{eqnarray}
\pami{where $R(S)$ is the score as a function of a set of states $S=\left[ s_1, s_2,...,s_i,...,s_N \right]$, $i$ is the index of frame, $s_i$ is the state of the $i^{th}$ frame, the space of $s_i$ is $\{state1, state2, ...\}$, $N$ is the total number of frames, $\lambda$ is the weight that balances the potentials, and $u(.),b(.)$ are the unary and binary scoring functions, respectively.}

\noindent\textbf{Scoring functions.}
\pami{The unary scoring function is exactly the same as the scores in Sec.~\ref{sec.ustate}.
The binary scoring function is defined below,
\begin{eqnarray}
&& \textit{for } i\notin C\textit{; if } s_i\neq s_{i+1} \textit{, } \\
&& b(s_i,s_{i+1})=-\inf \textit{; otherwise, } b(s_i,s_{i+1})=0~,
\end{eqnarray}
which means no change is allowed when the $i^{th}$ frame is not a change candidate;
\begin{eqnarray}
&&\textit{for } i\in C\textit{; if } s_i\neq s_{i+1} \textit{, } \\
&&b(s_i,s_{i+1})=-S(\bar{f}_i,\bar{f}_{i+1}) \textit{; otherwise, }\\
&&b(s_i,s_{i+1})=S(\bar{f}_i,\bar{f}_{i+1})~,\nonumber
\end{eqnarray}
where $S(\bar{f}_i,\bar{f}_{i+1})$ is the cosine similarity between $\bar{f_i},\bar{f}_{i+1}$,
\hl{$\bar{f}_i$ is the average frame-based deep features between the change candidate immediately before the $i^{th}$ frame and the change candidate at the $i^{th}$ frame, and $\bar{f}_{i+1}$ is the average frame-based deep features between the change candidate at the $i^{th}$ frame and the change candidate immediately after the $i^{th}$ frame.}
We apply a dynamic programming inference procedure to predict the states maximizing $R(S)$.}

\begin{figure*}[t]
\centering
\includegraphics[width=1\textwidth]{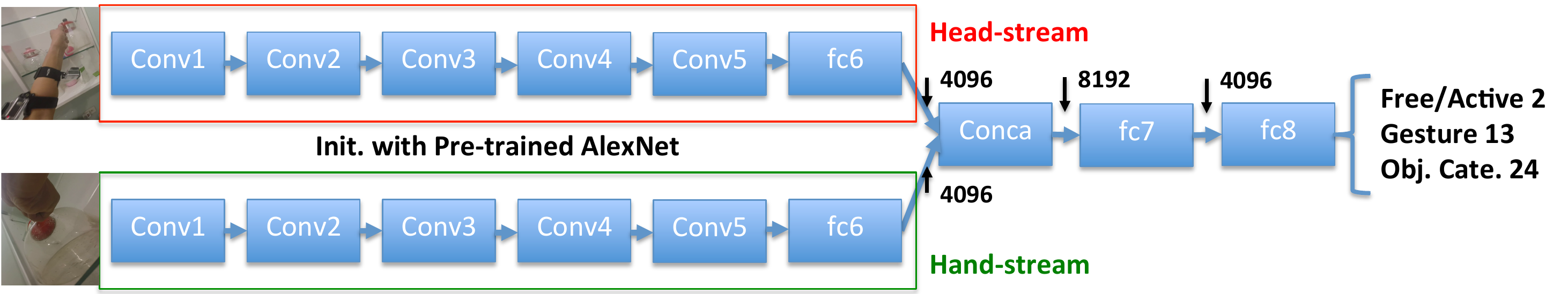}
  \caption{\small Architecture of our two-streams CNN. The top and bottom streams take the HeadCam and the HandCam, respectively, as inputs.
  The two-streams CNN is used to predict Free/Active states, 13 hand gesture states, or 24 object category states. Conv, fc, and Conca denote convolution, fully-connected, and concatenate, respectively.}
  \label{fig:2sCNN}
\end{figure*}
\cutsubsectionup
\subsection{Deep Feature}\label{sec.Dmodel}\cutsubsectiondown
\pami{We extract our deep feature from the well-known AlexNet~\cite{NIPS2012Alex} CNN model.
Instead of using the pre-trained $1K$ dimension final output as feature,
we try different design choices to address the following questions:
(1) which layer should we extract feature? and (2) will finetuning improve recognition from the new HandCam observation?
In our pilot experiment, we found that a compact six layers model achieves the best accuracy, while being more computationally efficient than the original AlexNet (see Appendix~\ref{sec.apDF}).
Hence, we use the fc6 output of AlexNet by default in all experiments of this paper. 
In Sec.~\ref{sec.Exp}, we also show that finetuning consistently improves state prediction accuracy.}

\cutsubsectionup
\subsection{Object Discovery}\label{sec.DC}\cutsubsectiondown
Given many observation of how users interact with objects in a new scene, we propose a simple method to discover common object categories. Firstly, we predict the active hand segments which is typically over-segmented. Then, we calculate segment-base feature $\bar{f}$ as the average of the frame-based features and apply a hierarchical clustering method using cosine similarity to group similar segments into clusters. 
By assuming that the same object category is manipulated by hands multiple-times in a similar way, two similar segments likely corresponds to the same object categories.
In Sec.~\ref{sec.OC}, we show that our HandCam system can discover categories more accurately than a HeadCam system.


\cutsubsectionup
\subsection{Combining HandCam with HeadCam}\label{sec.comb}
\cutsubsectiondown
Since our goal is to achieve the best accuracy, we would like to combine HandCam with HeadCam to do even better.
\pami{Intuitively, HeadCam should be complementary to HandCam in some ways.
Firstly, sometimes HandCam is occluded by other objects, whereas HeadCam keeps a clear view of the hands (Fig.~\ref{fig.3D}(a)) due to required hand-eye coordination.
Second, HeadCam observed more scene/place information which might be useful.
For instance, we have used the observation from HeadCam to reconstruct\footnote{We use visualsfm \cite{Changchang13DV} for reconstruction.} the scene as well as localize the HeadCam in the scene as shown in Fig.~\ref{fig.3D}(b).
It is possible that certain place information observed by HeadCam can be beneficial for recognizing some hand states.}

We propose two approaches to combine HeadCam and HandCam.

\noindent\textbf{Feature concatenation.}
We simply concatenate the separately finetuned HeadCam and HandCam features.
Then, we use the concatenate feature to train the state classifier and state change detector as described before.
Although this is a straight forward approach, we show in our experiment that it already outperforms other settings.

\noindent\textbf{Two-streams CNN.}
\pami{Inspired by \cite{simonyan2014two}, we treat each camera as one unique stream and design a novel two-streams CNN.
Each stream first feeds-forward through a CNN of six layers with the same first six layers in the AlexNet~\cite{NIPS2012Alex} architecture.
Then, the fc6 (each with 4096 dimension) outputs of both streams are concatenated (total 8192 dimension) before forwarding to the next two
fully connected layers. We use this two-streams CNN to predict Free/Active states, 13 hand gesture states, or 24 object category states.
Please see Fig.~\ref{fig:2sCNN} for the detail architecture.
The model weights of both streams are initialized with the ImageNet pre-trained AlexNet model.
Then, we finetune the full two-streams CNN with parameters detailed in Sec.~\ref{sec.dd}.
After finetuning, we take the last hidden representation of our two-streams CNN as feature to
train the state classifier and state change detector as described before.
Our experiment shows that jointly finetuning two-streams achieves the best performance in four out of five tasks.
}




\cutsectionup
\section{Dataset}\label{sec.D}
\cutsectiondown


\pami{We have collected a new synchronized ``HandCam" and ``HeadCam" video data for hand states recognition.} Our dataset contains $20$ round of data collection ($60$ video sequences), where each round consists of three synchronized video sequences (two from HandCam and one from HeadCam). In total, our dataset contains $\sim 115.5$ minutes of videos, which is at a similar scale of the egocentric video dataset \cite{damen2014you}.
\pami{For HandCam, we ask each user to mount the camera so that the palm is in the center of the camera.
For HeadCam, we found it is much harder to be mounted correctly by each user.
Hence, we help each user to mount the HeadCam so that the user's gaze is in the center of the camera while manipulating objects.
Our dataset will be publicly available. 
A subset of videos can be viewed in \url{https://www.youtube.com/watch?v=nPi5L-xG5Ng}.}

\begin{figure*}[!t]
\centering
\includegraphics[width=1\textwidth]{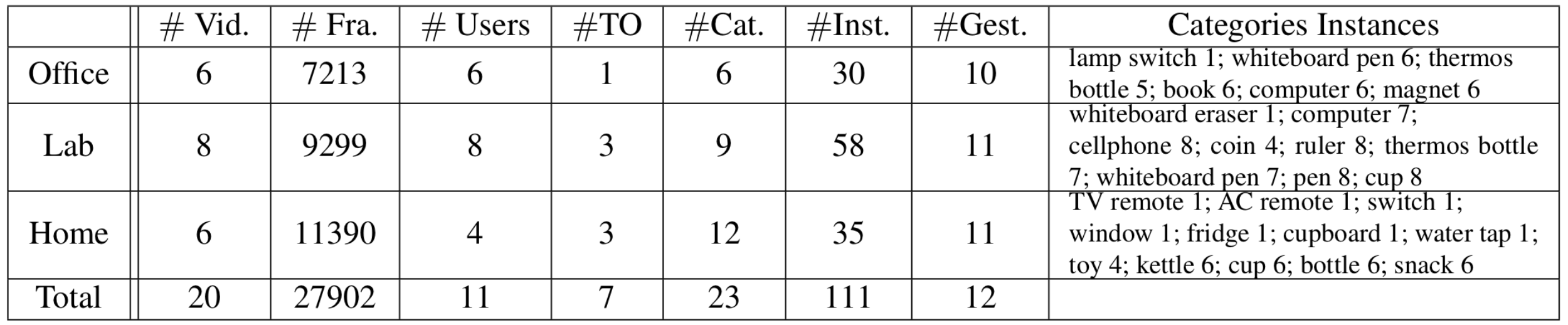}
  \caption{\small Statistics of our HandCam dataset. Vid. stands for videos. Fra. stands for frames. TO stands for task-order. Cats. stands for categories. Inst. stands for instances. Gest. stands for gestures.}
  \label{fig:stat}
\end{figure*}

In order to thoroughly analyze tasks involving recognizing object category, hand gesture, etc., we explicitly collect videos in multiple indoor scenes, interacting with multiple object categories, and multiple object instances within each category. Moreover, we ask multiple users to wear our system in a casual way to consider the variation introduced by multiple users.
A thorough statistics is shown in Table.~\ref{fig:stat}. We summarize the properties of our dataset below.
\begin{itemize}
\item Scene: We have collected videos in three scenes: a small office, a mid-size lab, and a large home (Fig.~\ref{fig.3D}(b)), where office and lab involve many similar object interactions, but involve very different object interactions as in home.
\item Task: We pre-define a number of tasks for users to follow. To increase variation, each user randomly selects a task-order to perform.
\item \pami{Object category: We prepare many instances for most object categories: typically movable objects such as toys, kettles, cups, bottles, cookies, coins, cellphones, rulers, pens, books, and magnets.  We ensure that these instances are separable in our train/test splits.}
\item User: We have 11 unique users involved in collecting the videos.
\end{itemize}


\begin{figure}[!t]\cuttableup
\centering
\includegraphics[width=0.5\textwidth]{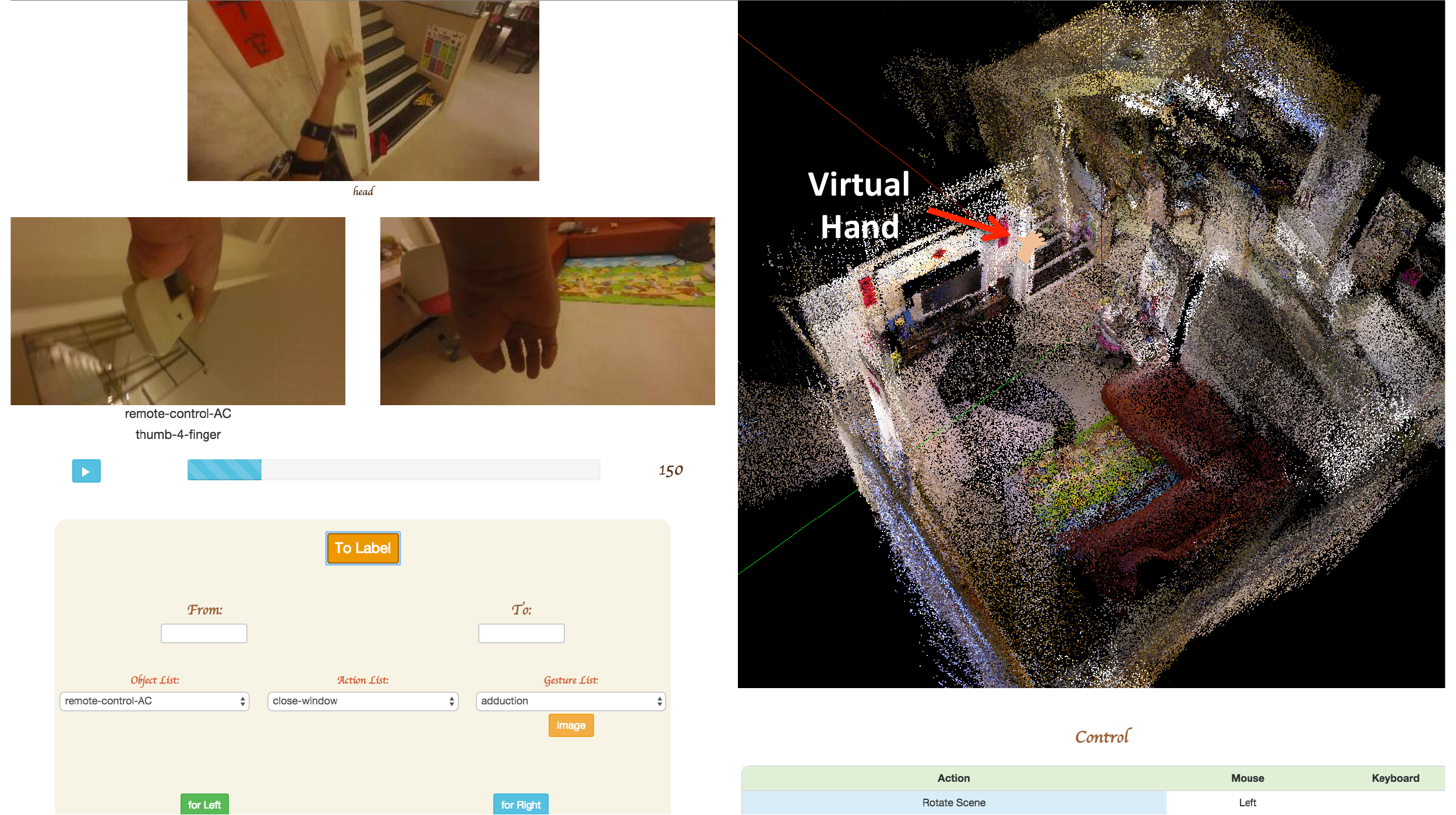}
 \cutcaptionup
  \caption{\small Snapshot of our data viewer for annotators to simultaneously watch 3 videos for labeling the hand states.}
  \label{fig:label}
  \cutcaptiondown
\end{figure}

\noindent\textbf{Annotating interface.}
\pami{For annotating hand states, we ask human annotators to simultaneously watch the synchronized three videos (i.e., two HandCams and one HeadCam) to make their label decision.
Moreover, the estimated 3D trajectory of the users in a 3D scene is also visualized beside the videos to help the annotator understand the situation.
A snapshot of our viewer is shown in Fig.~\ref{fig:label}.
}

\noindent\textbf{Training v.s. testing set.}
We have two settings.
Firstly, we train on office and lab. Then, we test on home. We refer this as ``Home" setting. This is a challenging setting, since home is an unseen scene and there are many unseen object categories\footnote{We evaluate object discovery performance in Home setting, since training and testing consists of different object categories.} and activities.
In the second setting, we evenly divide the video sequences into half for training and the remaining half for testing in all scenes. We refer this as ``AllScenes" setting.

\cutsectionup
\section{Implementation Details}\label{sec.dd}
\cutsectiondown
\noindent\textbf{Camera system.} Our system consists of three GoPro 3+ cameras to record videos with 1920x1080 resolution at 60fps and we process them at 6 fps. In order to record synchronized video sequences among two HandCams and one head-mounted camera, we use the GoPro Wi-Fi remote control to start and end recording all three cameras at the same time.

\noindent\textbf{Alignment.}
We achieve stable result by setting $\beta^{th}=40$ and trying seven scales (i.e., $\left[ 0.9,1,1.1,1.2,1.3,1.4,1.5\right]$) in our multi-scales alignment method.

\noindent\textbf{Training.}
We set SVM regularization parameters, parameter $d$ of state change features, and  $\lambda$ automatically using 5-fold cross-validation for each setting.
We finetune an imagenet pre-trained AlexNet on our dataset using the following parameters consistently for all tasks: maximum iterations = 40000, step-size = 10000, momentum = 0.9, every 10000 iteration weight decay =0.1, and learning rate = 0.001. To augment our dataset, we flip the left HandCam frames horizontally and jointly trained with the right HandCam frames.

\noindent\textbf{Training two-streams CNN.}
\pami{For finetuning the two-streams CNN on our dataset, we set
maximum iterations = 10000, step-size = 2500, momentum = 0.9, every 2500 iteration weight decay =0.1, and learning rate = 0.001.
We also augment our dataset by horizontal flipping frames.}

\cutsectionup
\section{Experiment Results}\label{sec.Exp}
\cutsectiondown


\begin{figure}[!t]\cuttableup
\centering
\includegraphics[width=0.48\textwidth]{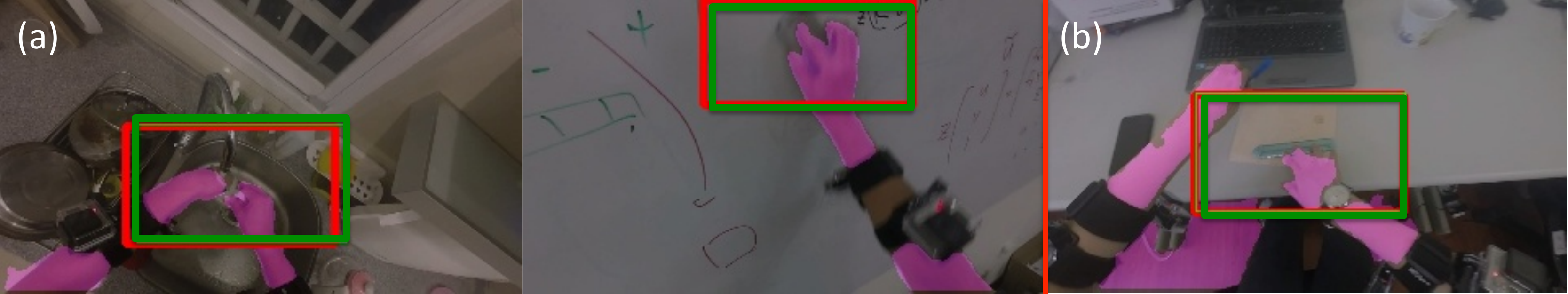}
\cutcaptionup
\caption{\small Typical hand segmentation (pink-mask) and manipulation region prediction (green-boxes). Panel (a) shows two good examples. The prediction and ground truth are shown in green boxes and red boxes, respectively. Panel (b) shows a typical hand mask mistake where the wooden floor with skin-like color is segmented as hands. When the mistake is not significant, the predicted manipulation region can still be correct.}\label{fig.7}
\cuttabledown
\end{figure}

We evaluate three state recognition tasks: free v.s. active, gesture, and object category recognition.
Some tasks are conducted in two train/test settings: “Home” and “AllScenes” as described in Sec.~\ref{sec.D}. In addition, a challenging object discovery task is evaluated in “Home” setting. All the following experiments are conducted using fc6 features in Alexnet. 

\noindent\textbf{HeadCam baseline.}
We apply state-of-the-art hand segmentation method \cite{li2013model} to predict manipulation region at each frame.
Similar to \cite{li2013learning}, we predict at most two boxes, one for left and one for right hands. Typical ground truth and predicted boxes are shown in Fig.~\ref{fig.7}.
Next, we crop the HeadCam images with respect to the predict manipulation region and apply the same methods introduced in this paper to recognize hand states  (see Fig.~\ref{fig.7}).
We also use improved dense trajectory~\cite{wang:2011} of the whole frame to capture the motion cues as a strong but time-consuming baseline.

\noindent\textbf{Method abbreviation.}
\hl{To faciliate discussion, we introduce the following abbreviations for different methods
\begin{itemize}
\item{HeadCam: IDT, BL, BLCrop, and BLCropFT.}\\
IDT is a HeadCam baseline using improved dense trajectory~\cite{wang:2011}.
BL is a HeadCam baseline without cropping image using pre-trained feature. BLCrop is a HeadCam baseline with manipulation region crop using
pre-trained feature. BLCropFT is BLCrop using finetuned feature.
\item{HandCam: NoAlign, Align, and AlignFT.}\\
 NoAlign is HandCam without hand alignment using pre-trained feature.
Align is HandCam with alignment using pre-trained feature. AlignFT is Align using finetuned feature.
\end{itemize}
}

\begin{figure}[!t]\cuttableup
\centering
\includegraphics[width=0.48\textwidth]{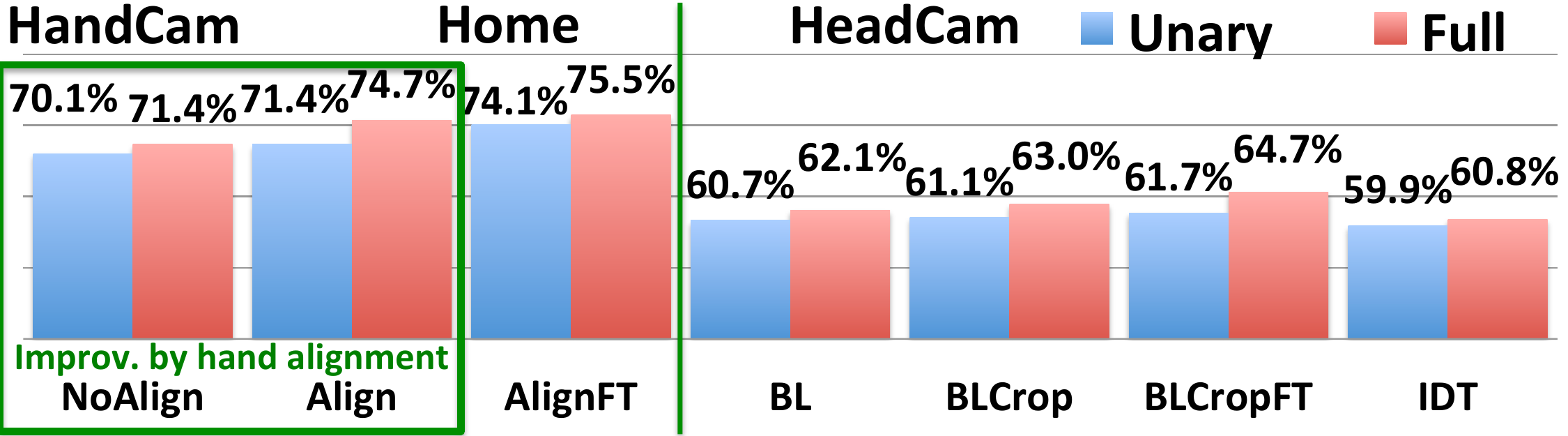}
\includegraphics[width=0.48\textwidth]{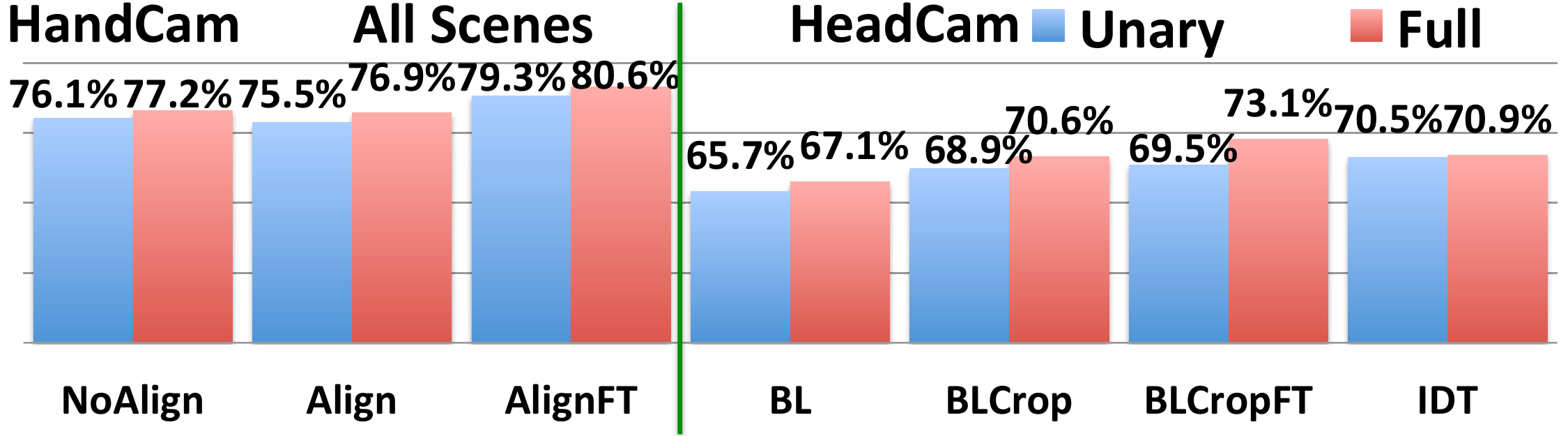}
\cutcaptionup\vspace{4mm}
\caption{\small Free v.s. active recognition accuracy. Left-panel shows the global accuracy in Home. Right-panel shows the global accuracy in AllScenes. In all plots, blue bar denotes frame-based classification relying on unary scoring function only. Red bar denotes our full model.}\label{fig.8}
\cuttabledown
\end{figure}

\cutsubsectionup
\subsection{Free v.s. Active Recognition}\label{sec.FAexp}
\cutsubsectiondown

Free v.s. active recognition accuracy (acc.) comparison
is shown in Fig.~\ref{fig.8}, where left-panel shows the acc. in Home and right-panel shows acc. in AllScenes.

\noindent\textbf{Pre-trained CNN.}
Using pre-trained CNN feature, our full method ($74.7\%$ acc. in Home and $76.9\%$ acc. in AllScenes)
is already consistently better than both the non-cropped ($62.1\%$ acc. in Home and $67.1\%$ acc. in AllScenes),
cropped ($63\%$ acc. in Home and $70.6\%$ acc. in AllScenes), and IDT HeadCam baselines.
These results confirm that HandCam is a great alternative to HeadCam systems.

\noindent\textbf{Unary v.s. Full.}
Our full model also consistently outperforms frame-based unary model in all settings and for
both HandCam and HeadCam (with the best improvement of $3.3\%$ in Home and aligned HandCam setting).

\noindent\textbf{Hand Alignment.}
Although in AllScenes, hand alignment shows no critical improvement. However, in the challenging Home setting, we confirm that hand alignment improves
acc. by $3.3\%$ from $71.4\%$ to $74.7\%$ (full model+pre-trained feature in green box of Fig.~\ref{fig.8}). This implies that hand alignment might be critical in practical across-scenes usage.

\noindent\textbf{Finetune CNN.}
Finetuning CNN shows consistent improvement in all settings and for both HandCam and HeadCam. Our finetuned full method achieves $75.5\%$ acc. in Home and $80.6\%$ acc. in AllScenes which is $10.8\%$ and
$7.4\%$ better than the finetuned cropped baseline.

\begin{figure}[!b]\cuttableup
\centering
\includegraphics[width=0.48\textwidth]{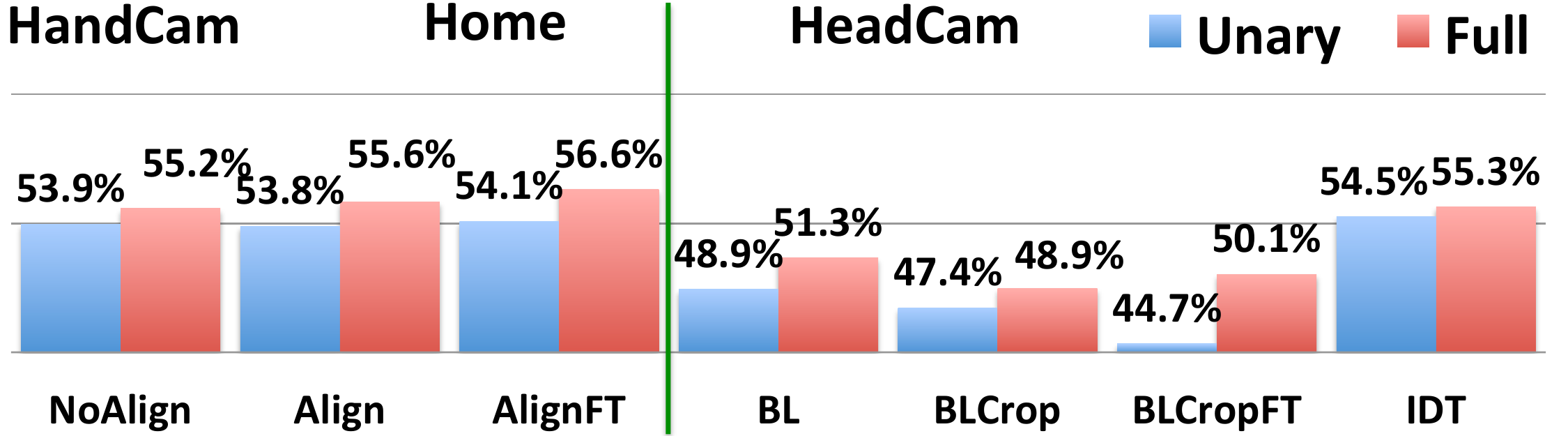}
\includegraphics[width=0.48\textwidth]{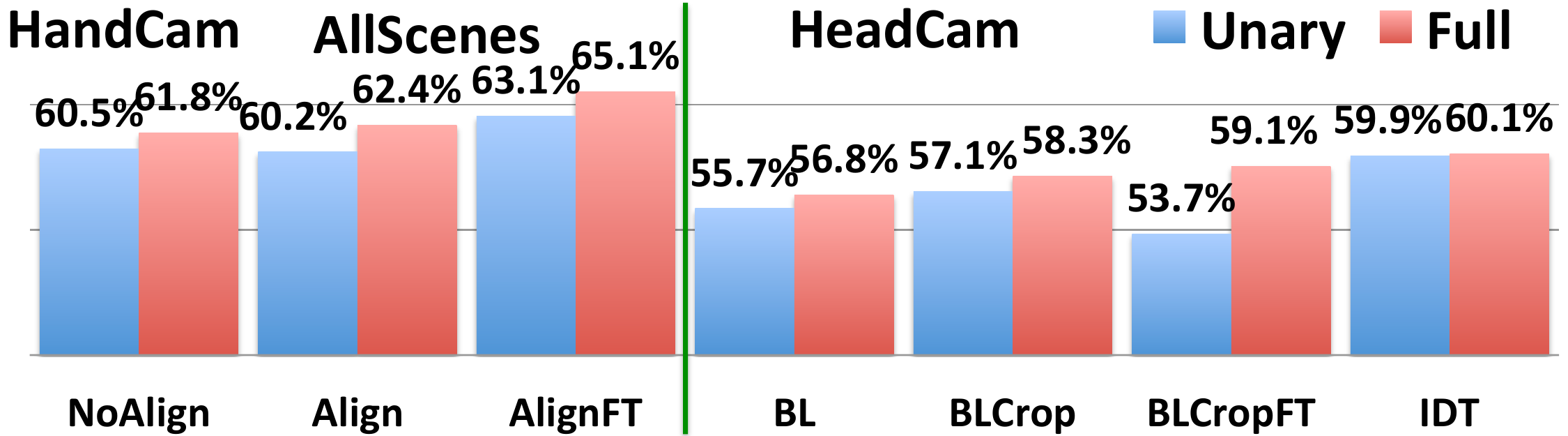}
\cutcaptionup\vspace{4mm}
\caption{\small Gesture recognition accuracy. Left-panel shows the global accuracy in Home. Right-panel shows the global accuracy in AllScenes. In all plots, red and blue bars are defined in Fig.~\ref{fig.8}}\label{fig.9}
\cuttabledown
\end{figure}


\cutsubsectionup
\subsection{Gesture Recognition}
\cutsubsectiondown

Gesture recognition accuracy comparison is shown in Fig.~\ref{fig.9}. Gesture recognition is more challenging than free
v.s. active recognition, since it needs to classify 12 + 1(free hand state) gestures. However, gesture recognition shares
the same trend in free v.s. active recognition. Except that, in Home setting, hand alignment only shows $0.4\%$ acc. improvement using pre-trained feature. Nevertheless, our finetuned full method achieves $56.6\%$ acc. in Home and $65.1\%$ acc. in AllScenes.
They are $6.6\%$ (Home) and $6\%$ (AllScenes) better than the finetuned cropped baseline,
and $1.3\%$ (Home) and $5\%$ (AllScenes) better than IDT.
This implies that IDT helps focus on fine gesture details in HeadCam.

\cutsubsectionup
\subsection{Object Category Recognition}\label{sec.OC}
\cutsubsectiondown
Object category recognition accuracy comparison in
AllScenes setting is shown in Fig.~\ref{fig.13}-Left. 
We found that it shares the same trend in Sec.~\ref{sec.FAexp}. 
Most importantly, our finetuned full method achieves
$66.5\%$ acc. which is $9.1\%$ and $7.9\%$ better than the finetuned cropped baseline and IDT baseline, respectively. 
This also implies that IDT helps focus on fine object details in HeadCam.

In Home setting, since many object categories are not observed in training, we evaluate the following object discovery task.
\begin{figure}[!t]\cuttableup
\centering
\includegraphics[width=0.48\textwidth]{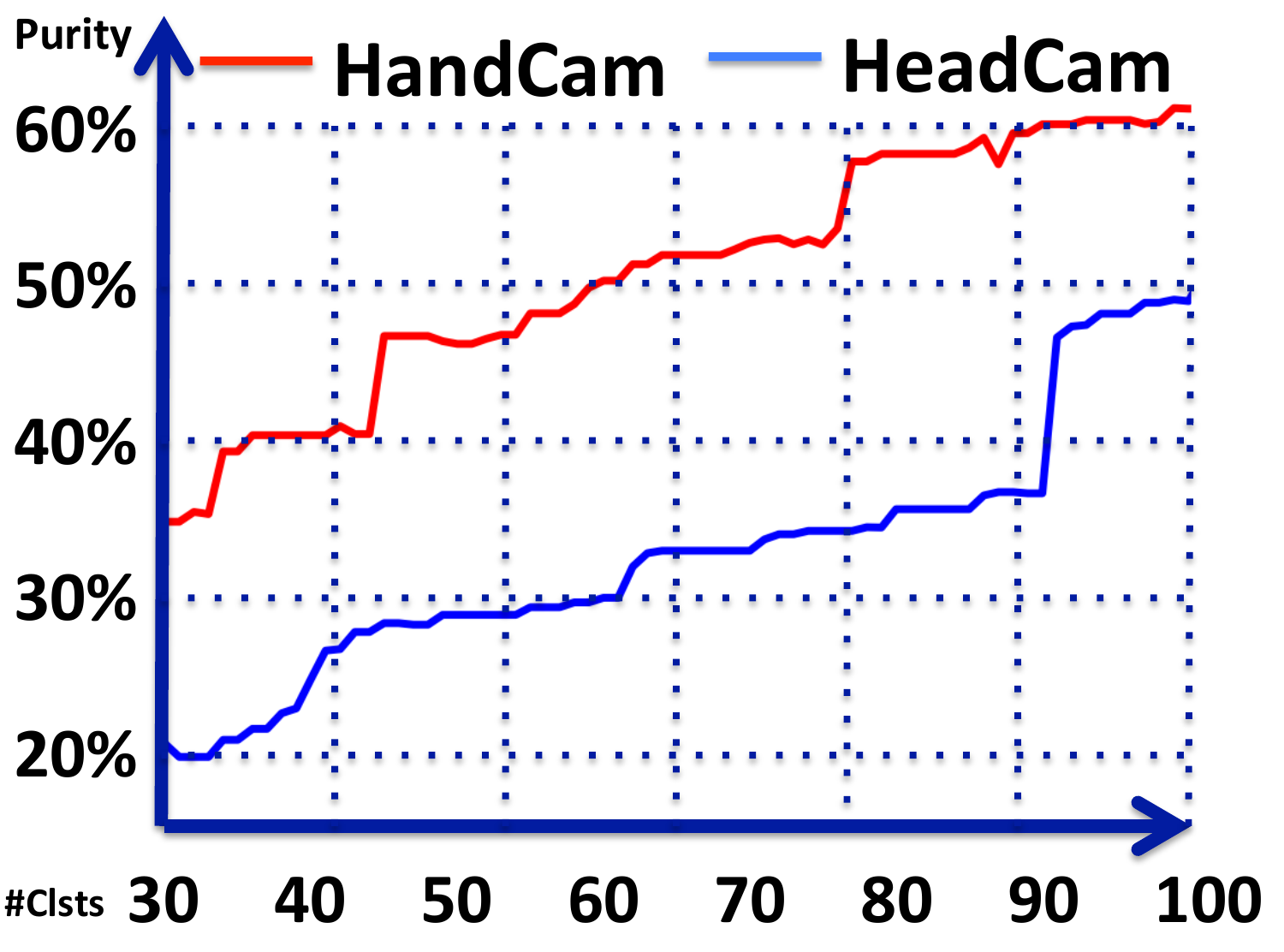}
\cutcaptionup
\caption{\small Purity for Object Category Discovery. Our HandCam (red curve) consistently ourtperforms HeadCam (blue curve) by about $10\%$ from 30 to 100 clusters.}\label{fig.11}
\cuttabledown
\end{figure}

\begin{figure}[!t]\cuttableup
\centering
\includegraphics[width=0.48\textwidth]{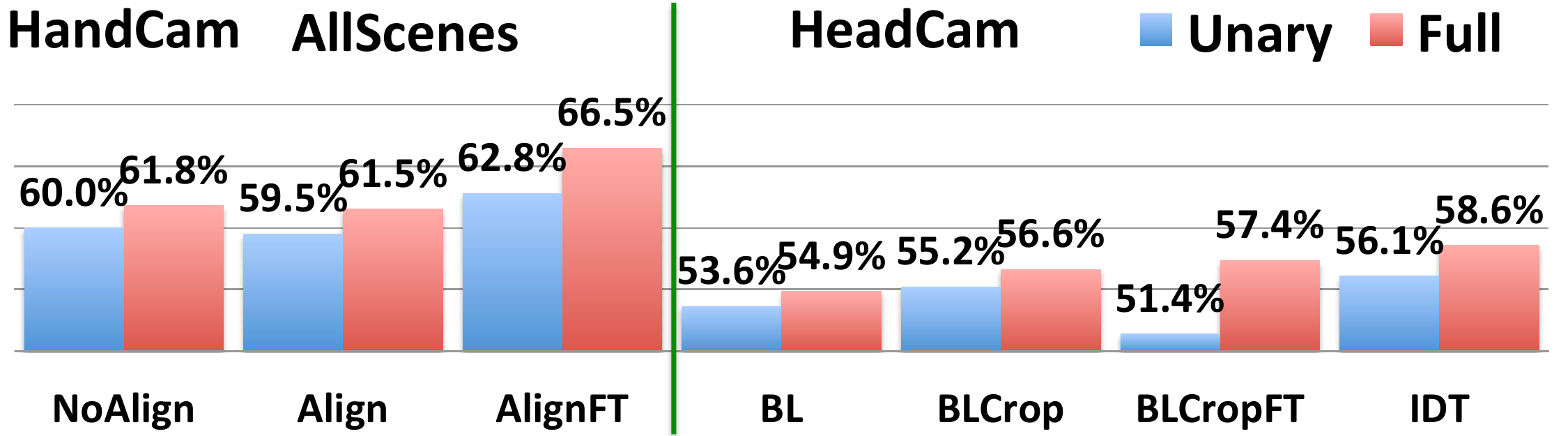}
\includegraphics[width=0.48\textwidth]{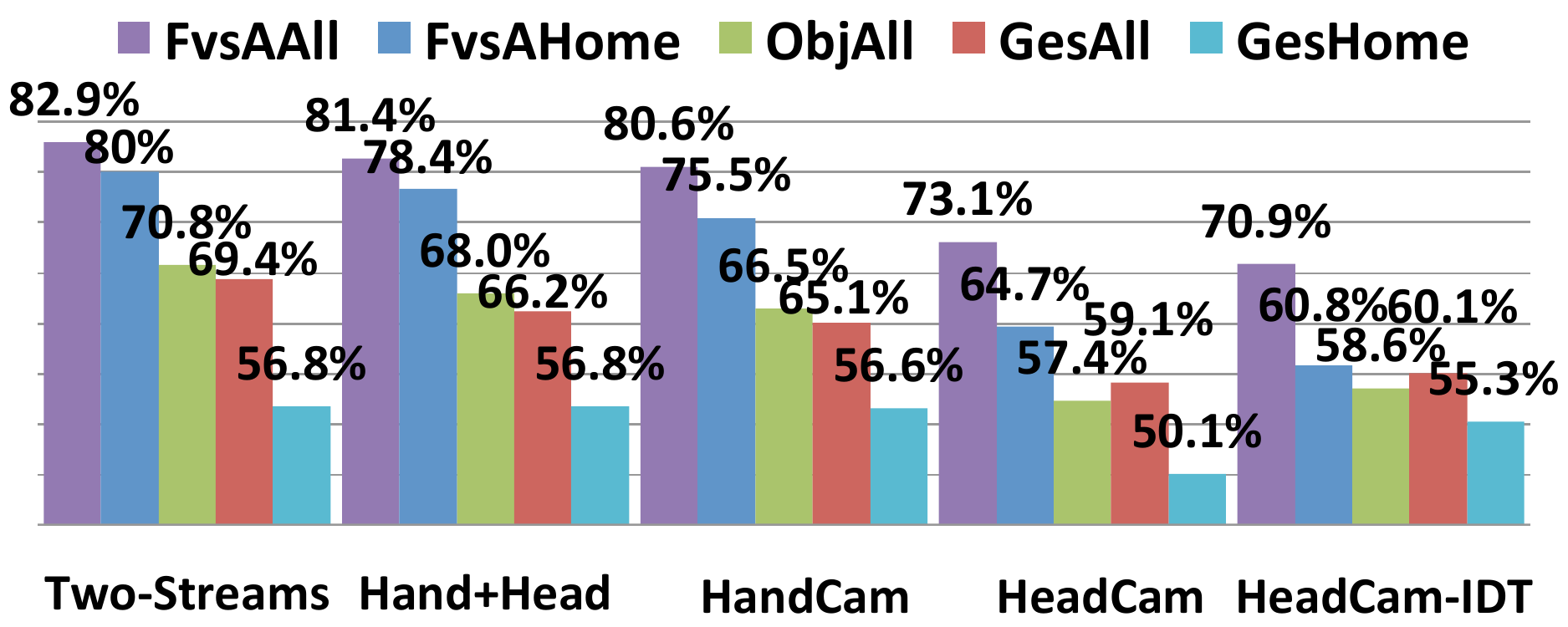}
\cutcaptionup\vspace{4mm}
\caption{\small 
Left panel: Object category classification comparisons in AllScenes. In all plots, red and blue bars are defined in Fig.~\ref{fig.8}.
Right panel: Comparing two-streams CNN with HandCam+HeadCam (finetuned feature concatenation), HandCam (AlignFT), HeadCam (BLCropFT), and HeadCam-IDT in five tasks: FvsAAll and FvsAHome stand for free v.s. active recognition in AllScenes and Home, respevtively;
ObjAll stands for object category recognition in AllScenes;
GesAll and GesHome stand for gesture recognition in AllScenes and Home, respectively.}\label{fig.13}
\cuttabledown
\end{figure}

\noindent\textbf{Category Discovery.}
We treat object category discovery as a clustering task, and compare the clustering results between our best HandCam configuration (i.e., AlignFT for
free v.s. active recognition) and the best HeadCam configuration (i.e., BLCropFT for free v.s. active recognition). We
report a modified purity (similarly defined in~\cite{Purity}) to focus
on discovering object categories (not “free-hand”) as described below. For all frames in each cluster, we obtain their
ground truth object labels and calculate the dominate label.
If the dominate label is not “hand-free” state, we count the number of discovered frames as the number of ground truth
labels equals the dominate label. This number is accumulated across all clusters. Then, the purity is the total number
of discovered frames divided by the number of true “active-hand” frames. We calculate the purity with different number of clusters (see Fig.~\ref{fig.11}),
and find that HandCam outperforms HeadCam by about $10\%$ from 30 to 100 clusters.



\begin{figure*}[!t]\cuttableup
\centering
\includegraphics[width=0.97\textwidth]{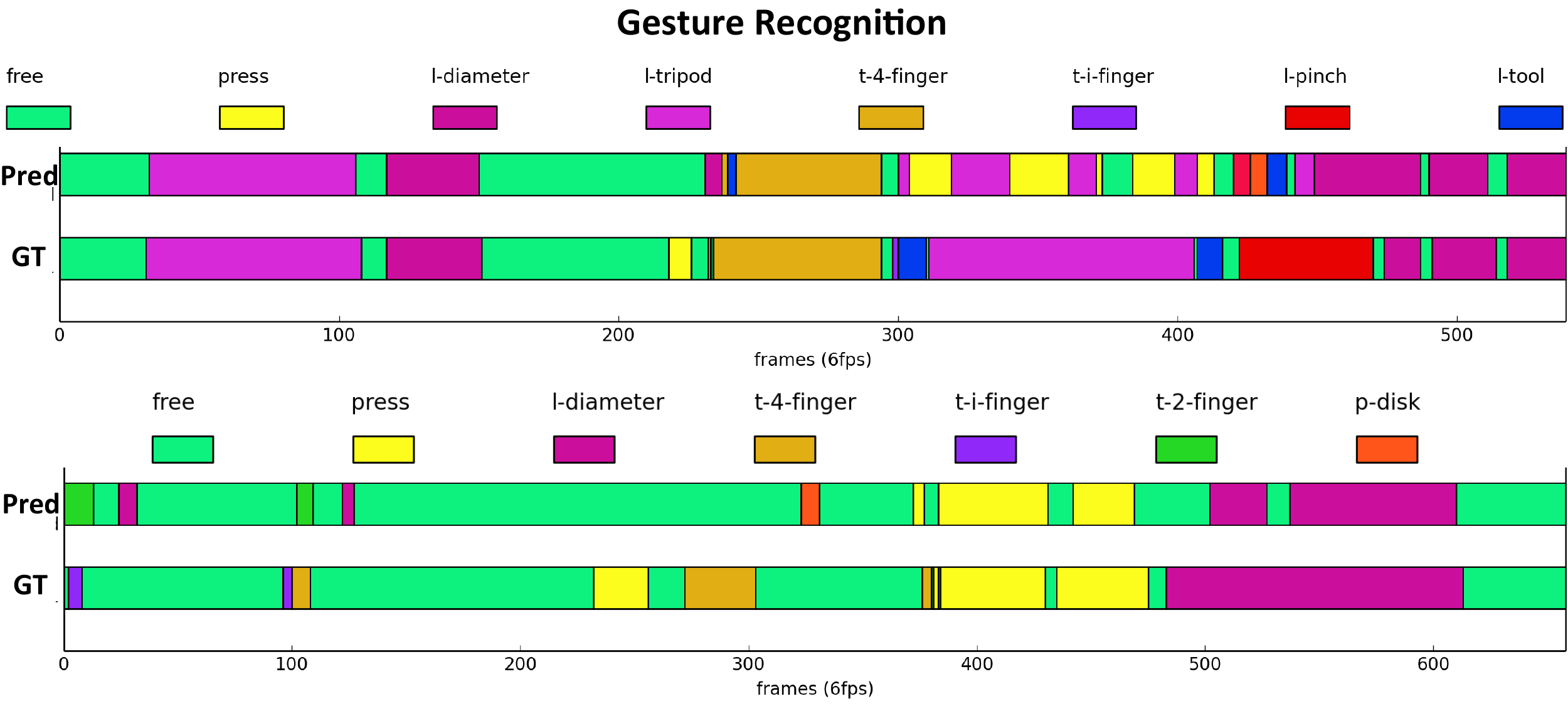}
\includegraphics[width=0.97\textwidth]{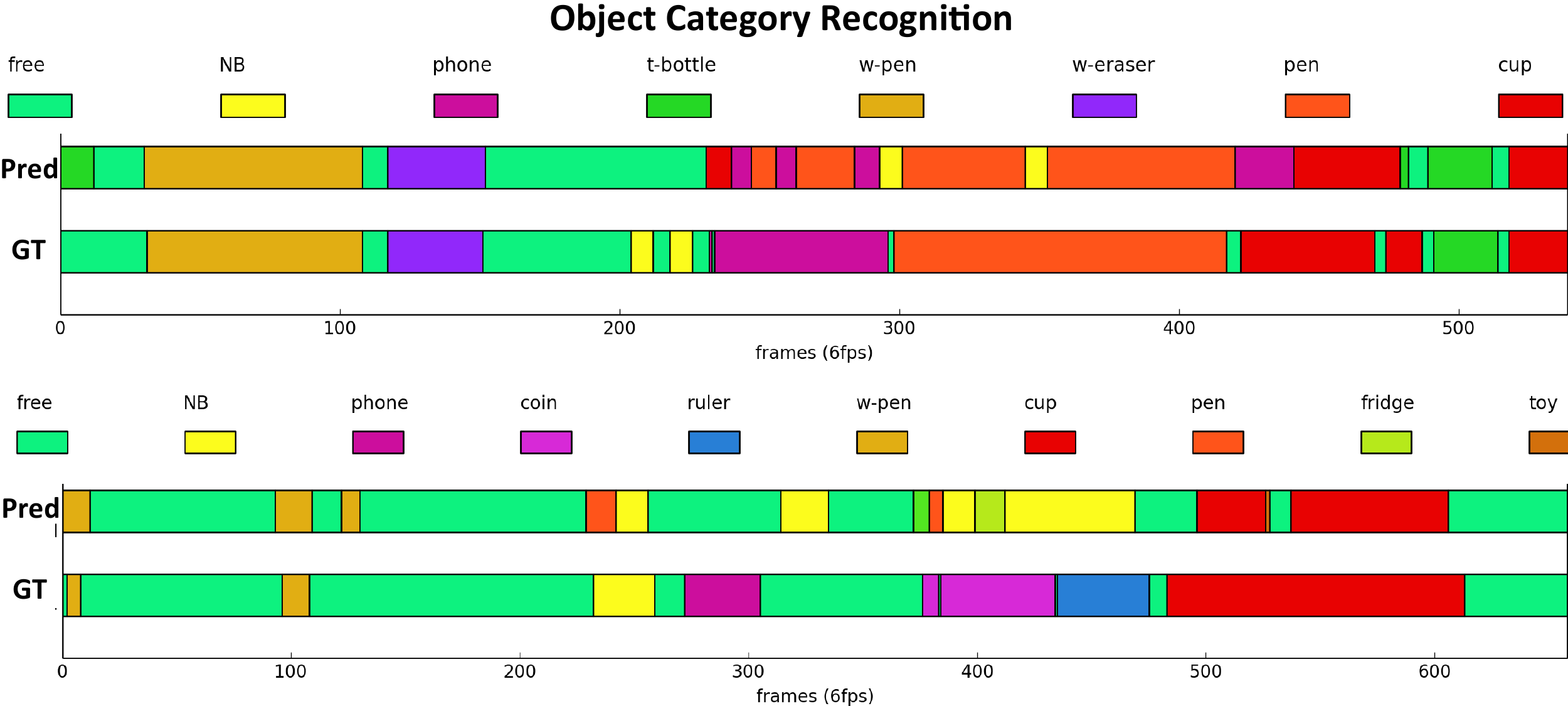}
\cutcaptionup\vspace{4mm}
\caption{\small Temporal visualization of predicted hand gesture (top-2-rows) and object category (bottom-2-rows) states from HandCam in AllScenes setting.
Pred stands for prediction of our two-streams CNN method.
The color-code of gesture (top-row) and object category (bottom-row) states are on top of each visualization.}\label{fig.vO}
\cuttabledown
\end{figure*}


\cutsubsectionup
\subsection{Combining HandCam with HeadCam}
\cutsubsectiondown
We show comparison among HeadCam best deep-config. (BLCropFT) and HeadCam motion (HeadCam-IDT), HandCam best
config. (AlignFT), and HandCam+HeadCam (feature concatenation), and our novel two-streams CNN. 
\pami{
All results are shown in Fig.~\ref{fig.13}-Right.
HandCam+HeadCam with simple feature concatenation already outperforms the HeadCam and HandCam separately consistently in all five tasks.
Most importantly, our novel two-streams CNN achieves the best performance in four our of five tasks (except gesture in Home setting).}

\begin{figure*}[!t]\cuttableup\vspace{-4mm}
\centering
\includegraphics[width=0.97\textwidth]{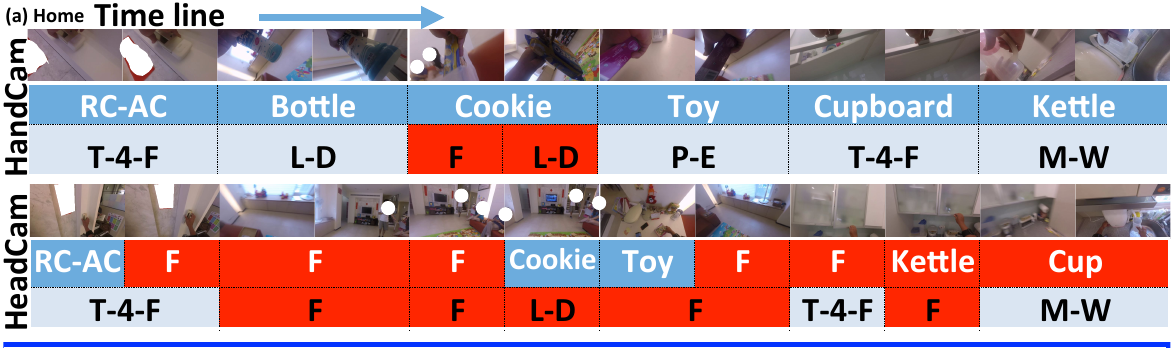}
\includegraphics[width=0.97\textwidth]{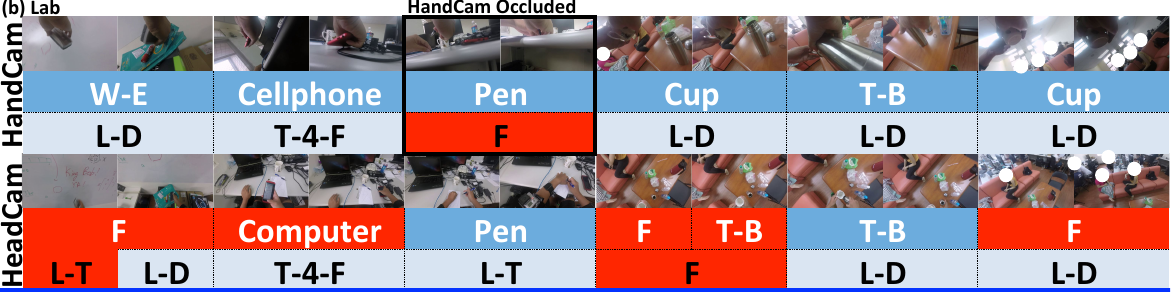}
\cutcaptionup
\caption{\small Typical examples of predicted object categories (white font) and gestures (black font). Panel (a,b) show the comparison
between HandCam (top-row) and HandCam (bottom-row) in home and lab scenes, respectively. Prediction errors are highlighted
in red background. T-B is thermos-bottle. M-W is medium-wrap. W-P is whiteboard-pen. W-E is whiteboard-eraser. RC-TV is remote-control-TV. RC-AC is remote-control-AC. L-D is large-diameter. L-T is lateral-tripod. T-4-F is thumb-4-finger. F is free. P-E is parallel-extension. We block a few regions in the images to avoid privacy concerns. HeadCam makes many more wrong prediction due to the variation of hand locations. Our HandCam sometimes makes wrong prediction when the camera is occluded.}\label{fig.12}
\cuttabledown
\end{figure*}
\cutsubsectionup
\subsection{Qualitative Results}
\cutsubsectiondown
\pami{We show temporal visualization of predicted v.s. ground truth hand states.
Fig.~\ref{fig.vO} shows the visualization of gestures states and object category states, respectively, in AllScenes setting of our two-streams CNN method.
Note that our method makes predict on all frames sampled at 6 fps, whereas the other recent method~\cite{OhnishiKKH15Wrist} makes prediction on pre-segmented active segments.
}

\section{Conclusion}
\cutsectiondown
We revisit HandCam (i.e., a wrist-mounted camera) system for recognizing various hand states.
To evaluate our system, we collect a new dataset with synchronized HandCam and HeadCam observing multiple object categories, instances, gestures in multiple scenes. HandCam with deep-learning-based method consistently outperforms HeadCam systems in all tasks by at most $10.8\%$ improvement in accuracy. We also observe that finetuning CNN consistently improves accuracy (at most $4.9\%$ acc. improvement). 
Most importantly, we show that combining HandCam with HeadCam using a novel two-streams CNN gives the best performance in four out of five tasks.
In the future, we will focus on two directions:
(1) studying how to improve the joint design of HandCam and HeadCam system;
(2) improving the system to collect a hourly long video per round of data collection to increase the scale of the dataset.
With more data and more sophisticated CNN architecture combining HandCam and HeadCam, we believe the recognition performance of our system can be greatly improved in the future.

\begin{appendices}

\section*{Appendices}

\section{Deep Feature}\label{sec.apDF}
\begin{figure}[!t]
\centering
\includegraphics[width=0.48\textwidth]{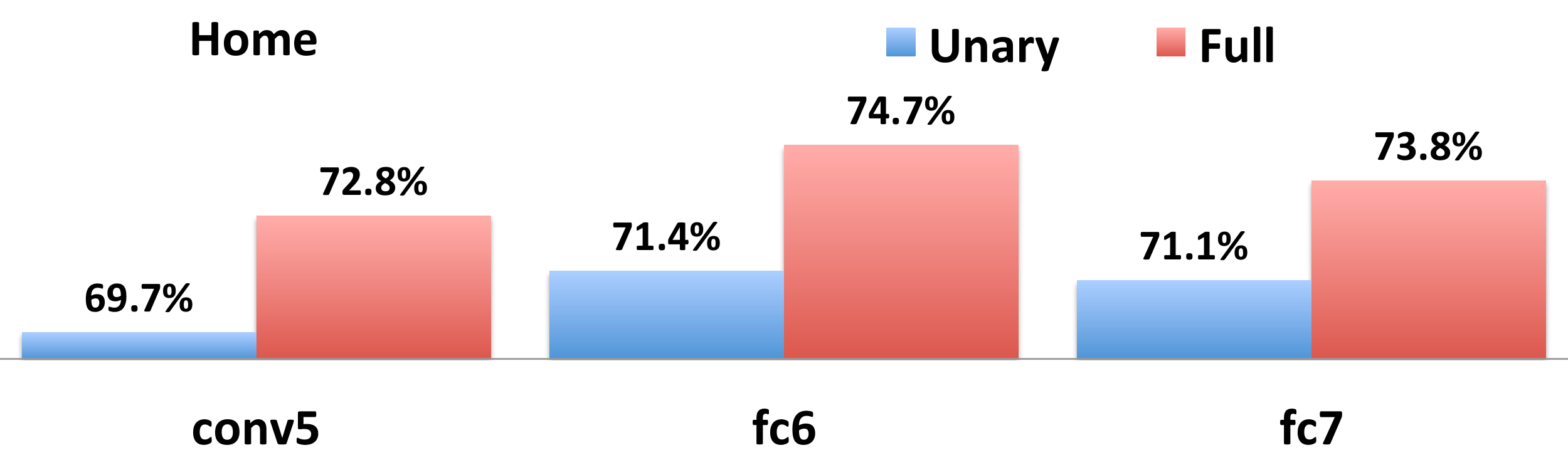}
\includegraphics[width=0.48\textwidth]{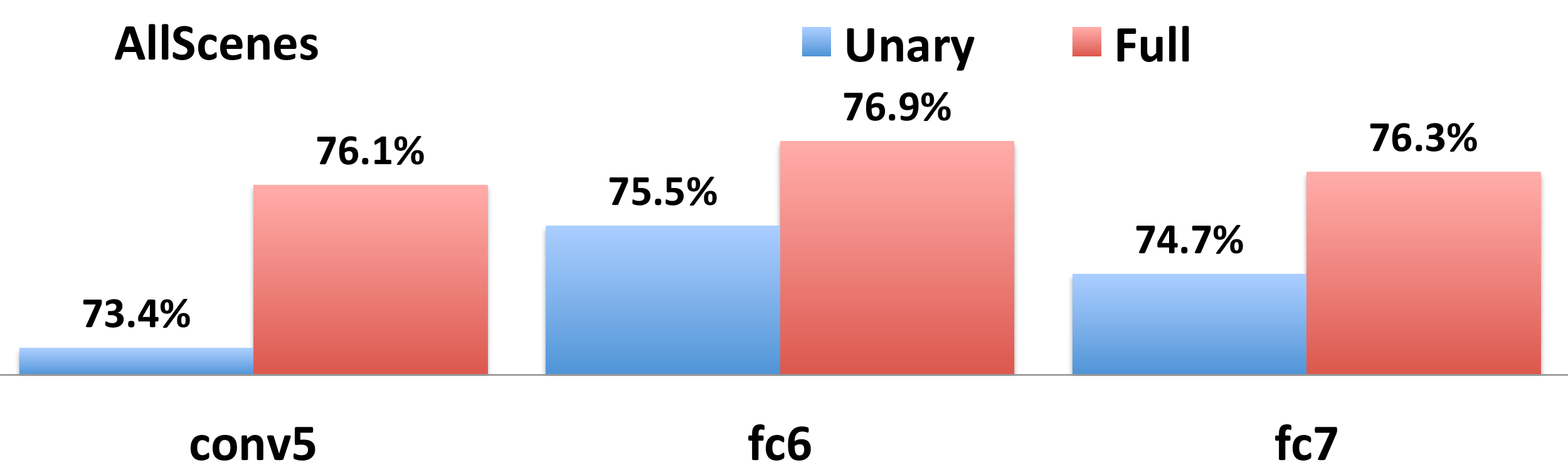}
\caption{\small Free v.s. active recognition accuracy comparison among three CNN architectures: conv5, fc6, and fc7. Top-row shows the global accuracy in Home. Bottom-row shows the global accuracy in AllScenes. In all plots, blue bar denotes frame-based classification. Red bar denotes our full model.}\label{fig.deep}
\end{figure}

We train three different CNN architectures: a 5-layers convolution model (conv5),  a 6-layers with last fully-connected layer (fc6), and a 7 layers with last 2 fully-connected layers (fc7)\footnote{All architectures have an additional softmax layer at the end.}. Their performances on free v.s. active recognition in Home and AllScenes settings using unary and our full models are show in Fig.~\ref{fig.deep}.
Our results show that a compact six layers model (fc6) achieves the best accuracy, while being more computationally efficient than the original Alexnet~\cite{NIPS2012Alex}.

\section{HeadCam Baseline}\label{sec.apHCBL}
 \begin{figure}[!t]
\centering
\includegraphics[width=0.48\textwidth]{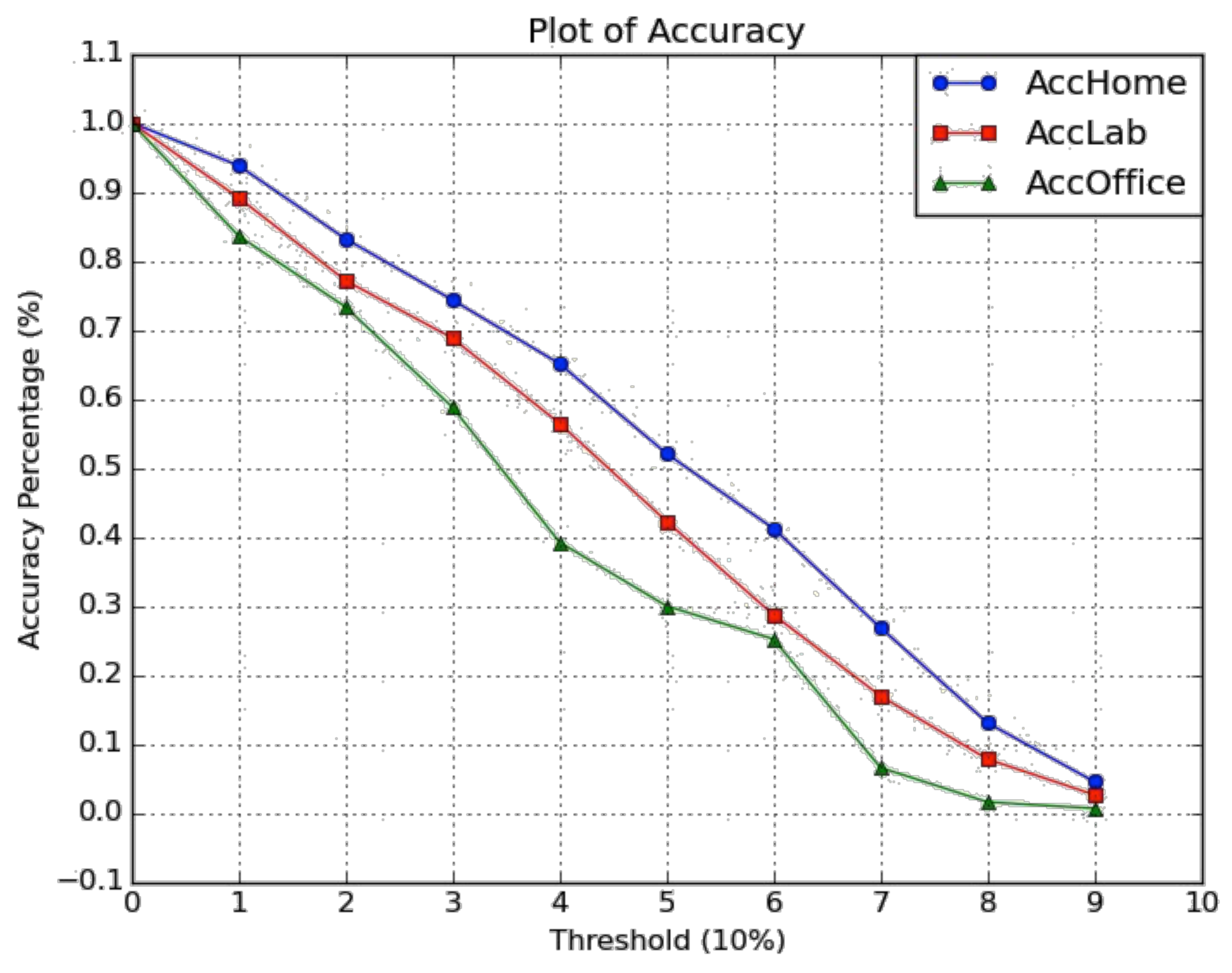}
\caption{\small Manipulation region prediction performance in Average Precision (Y-axis) with different overlap threshold (X-axis) for three scenes: home, office, and lab.}\label{fig.head}
\end{figure}
We first test the pre-trained hand segmenter~\cite{li2013model} but find the pre-trained model does not generalize well to images in our dataset.
In order to build a strong HeadCam baseline, we label the hand masks in one video per scene and retrain the hand segmenter.
We take a similar approach as in \cite{li2013learning} to predict the manipulation region (i.e., a bounding box) of each hand (i.e., both left and right hands) at each frame. In detail, we use a 10 nearest neighbor approach to retrieve the top 10 frames in our database with similar hand mask as the query hand mask.
Given the top 10 frames and their corresponding ground truth manipulation regions (i.e., boxes containing hands and objects), we use mean-shift to find the mode of the manipulation region, and use the mode region as the predicted region. In order to build a strong baseline, for each query frame, we use all frames of other videos in the same scene as frames in our database. We treat the manipulation region prediction task as an object detection task, and measure the overlap between the predicted region and ground truth region. We consider the prediction to be correct, if the overlap is more than a threshold.
In Fig.~\ref{fig.head}, we evaluate the percentage of correct prediction of the detection task at different thresholds. Our results show that manipulation region prediction is not perfect, even we have intentionally tried to improve its performance.
\end{appendices}

{\small
\bibliographystyle{ieee}
\bibliography{egbib2}
}

\end{document}